\documentclass[%
 aip,
 amsmath,amssymb,
 reprint,%
]{revtex4-1}

\usepackage{graphicx}
\usepackage{dcolumn}
\usepackage{bm}

\usepackage[utf8]{inputenc}
\usepackage[T1]{fontenc}
\usepackage{mathptmx}
\usepackage{etoolbox}
\usepackage{xcolor}

\makeatletter
\def\@email#1#2{%
 \endgroup
 \patchcmd{\titleblock@produce}
  {\frontmatter@RRAPformat}
  {\frontmatter@RRAPformat{\produce@RRAP{*#1\href{mailto:#2}{#2}}}\frontmatter@RRAPformat}
  {}{}
}%
\makeatother
\begin{document}

\preprint{AIP/123-QED}

\title[Noise mitigation strategies in physical FNNs]{Noise mitigation strategies in physical feedforward neural networks}
\author{N. Semenova}
 \email{semenovani@sgu.ru}
\affiliation{Département d’Optique P. M. Duffieux, Institut FEMTO-ST, Université Bourgogne-Franche-Comté CNRS UMR 6174, Besançon, France}%
\affiliation{Institute of Physics, Saratov State University, 83 Astrakhanskaya str., 410012 Saratov, Russia}
\author{D. Brunner}
 \email{daniel.brunner@femto-st.fr}
\affiliation{Département d’Optique P. M. Duffieux, Institut FEMTO-ST, Université Bourgogne-Franche-Comté CNRS UMR 6174, Besançon, France}%

\date{\today}

\begin{abstract}

Physical neural networks are promising candidates for next generation artificial intelligence hardware.
 In such architectures, neurons and connections are physically realized and do not leverage digital concepts with their practically infinite signal-to-noise ratio to encode, transduce and transform information.
 They therefore are prone to noise with a variety of statistical and architectural properties, and effective strategies leveraging network-inherent assets to mitigate noise in an hardware-efficient manner are important in the pursuit of next generation neural network hardware.
 Based on analytical derivations, we here introduce and analyse a variety of different noise-mitigation approaches.
 We analytically show that intra-layer connections in which the connection matrix's squared mean exceeds the mean of its square fully suppresses uncorrelated noise.
 We go beyond and develop two synergistic strategies for noise that is uncorrelated and correlated across populations of neurons.
 First, we introduce the concept of \emph{ghost neurons}, where each group of neurons perturbed by correlated noise has a negative connection to a single neuron, yet without receiving any input information.
 Secondly, we show that pooling of neuron populations is an efficient approach to suppress uncorrelated noise.
 As such, we developed a general noise mitigation strategy leveraging the statistical properties of the different noise terms most relevant in analogue hardware. 
 Finally, we demonstrate the effectiveness of this combined approach for trained neural network classifying the MNIST handwritten digits, for which we achieve a 4-fold improvement of the output signal-to-noise ratio and increase the classification accuracy almost to the level of the noise-free network.

\end{abstract}

\maketitle

\section{Introduction}

During the past years, neural networks (NNs) have provided solutions to previously unsolvable computing problems \cite{Lecun2015}.
 Among others, these tasks include image recognition and classification \cite{Krizhevsky2017,Maturana2015}, improvement of sound recordings, speech recognition \cite{Graves2013} and prediction of climatic phenomena \cite{Kar2009}.
 The basic principle of NNs is signal propagation between nonlinear neurons along connections according to some connection coefficients or connection weights.
 Among the most pressing objectives today is to implement NN topologies in hardware that drastically reduces the energy consumption compared to current NN hardware, and research activity along these lines has lately exploded.
 Special purpose NN chips, i.e. the newest generation of tensor and graphic processing units, allow low (2-6 bit) resolution computing \cite{Gupta2015}.
 
Combined with the need for removing the von Neumann bottleneck, the interest into low precision digital NN computing actually suggest analogue implementations of NN, i.e. in-memory computing leveraging computing with physical neural networks \cite{Wright2022,Markovic2020}, as promising substrates.
 At current digital resolutions for NN computing, analogue implementations substantially profit from the favorable energy usage per unit of information given by fundamental thermodynamics \cite{Boahen2017}.
 Physical NNs target encoding a NN's topology in a tunable analogue circuit, for example in electronic \cite{Wang2018,Lin2020,Xia2019} and photonic systems \cite{Feldmann2021}.
 Physical NNs leveraging lasers \cite{Brunner2013a, Nguimdo2020, Huang2022, Wang2022, Panda2022}, and spin-torque oscillators \cite{Torrejon2017} as neurons have been demonstrated.
 A physical NN's connections have been realized using holography \cite{Psaltis1990}, diffraction \cite{Bueno2018, Lin2018}, integrated networks of Mach-Zender modulators \cite{Shen2017}, wavelength division multiplexing \cite{Tait2017}, and 3D printed optical interconnects \cite{Moughames2020,Dinc2020,Moughames2020a}. 
Such, analog NN hardware is fundamentally prone to noise, and previous works provide strategies for reducing an analogue physical neuron's noise specific for the particular hardware \cite{Dolenko1993,Misra2010,Dibazar2006,Soriano2015,Frye1991}.
 Previously, we derived analytical descriptions of noise propagation and potential accumulation in deep NNs \cite{Semenova2019,Semenova2022}.
 The analytic equations describing the signal to noise ratio (SNR) at the output of a physical NN identified the most relevant sources of noise as well as strategies for effective noise suppression.  
 Here, we introduce and discuss several approaches of noise mitigation that are tailored to mitigate the most relevant generic types of noise.
 Importantly, individual strategies can be combined into a general noise mitigation framework that is adjustable to the particularities of a specific NN hardware architecture.
 
First, we discuss which sections of NNs are most affected by particular noise types, which is followed by analytically describing how one can leverage statistical properties of a NNs connectivity matrices to reduce noise simply by means of a noise-optimized topology.
 Next, we go beyond pure statistics-based strategies and introduce  \emph{ghost neurons}.
 A ghost neuron is a single neuron per layer that does not receive any input, and whose output is subtracted from each neuron in this layer in order to remove correlated additive noise.
 Furthermore, we discuss the impact of pooling neuron populations within layers, i.e. combining several neurons receiving the same input into one 'macro' neuron.
 Averaging the outputs of its individual elements, the macro neuron has reduced sensitivity to both types of uncorrelated noise.
 Finally, we apply the suggested noise mitigation techniques to reduce noise in NN trained to recognize MNIST digits database, where we achieve an excellent 4-fold suppression of noise at the final output layer of the 3 layer NN.

\section{System under study}

\begin{figure}[h!]
\center{\includegraphics[width=\linewidth]{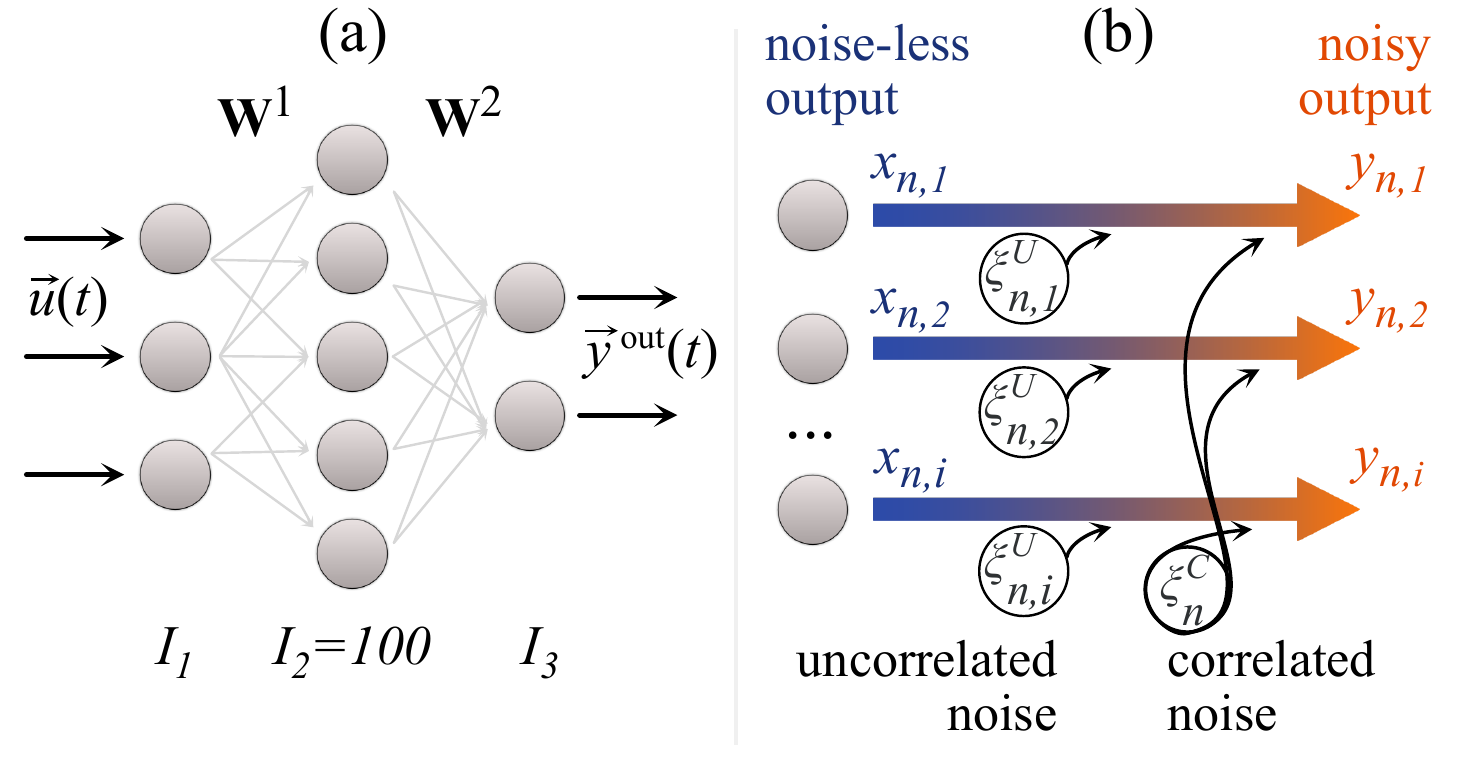}}
\caption[]{(a) Schematic representation of a feed-forward neural network, and (b) how uncorrelated and correlated noise is introduced into neurons.}\label{fig:FNN_scheme}
\end{figure}

Our work focuses on deep feed-forward neural networks (FNNs).
 These are networks consisting of a linear input and output layer, plus potentially several hidden layers, and information propagates strictly uni-directional from a preceding to a following layer.
 A schematic illustration of such a FNN is shown in Fig.~\ref{fig:FNN_scheme}(a).
 The input layer comprising $I_1$ linear neurons receives input according to vector $\vec{u}(t)$, while the output layer with $I_3$ linear or nonlinear neurons provides output vector $\vec{y}^\mathrm{\ out}(t)$.
 Here, we generally consider one hidden layer with $I_2=100$ neurons with $f(\cdot)$ as their nonlinear activation function.
 The connection topology between layers $n$ and $(n+1)$ is captured by connection matrix $\mathbf{W}^n$ that is of dimension $I_n\times I_{n+1}$.
 Then the signals coming to neurons belonging to layer $(n+1)$ are $\vec{a}_{n+1}$, and after activation function they transform to the noise-less signals $\vec{x}_{n+1}$:
\begin{equation}\label{eq:neuron_no_noise}
\vec{a}_{n+1}=\mathbf{W}^n\cdot\vec{y}_{n},\ \ \ \ \ \ \ \vec{x}_{n+1}=f(\vec{a}_{n+1}),
\end{equation}
where $\vec{y}_n$ is the noisy signal from layer $n$. If noise is turned off then $\vec{y}_n=\vec{x}_n$

Thus, we come to the main aspect of this article: the mitigation of noise and avoiding its accumulating as information propagates to the physical NN's output $\vec{y}^\mathrm{\ out}(t)$.
 Previously, we analytically captured the general impact of noise on FNNs with linear \cite{Semenova2019} and nonlinear neurons that were trained with with error back propagation \cite{Semenova2022}.
 Here, we substantially extend our analysis and derive noise reduction strategies.
 Here, noise is introduced identical as in \cite{Semenova2019,Semenova2022}, and we include additive and multiplicative noise, which are the most common types of noise found in analogue hardware.
 The signals of noisy neurons $i$ in layer $n$ are
\begin{equation}
\begin{array}{l}
y_{n,i}=x_{n,i}+\sqrt{2D_A}\cdot\xi^A_{n,i}(t) \ \ \text{additive noise},\\
y_{n,i}=x_{n,i}\cdot(1+\sqrt{2D_M}\cdot\xi^M_{n,i}(t)) \ \ \text{multiplicative noise},
\end{array}
\end{equation}
where  indices $A$ and $M$ indicate the noise type.
 $\xi$ is the white Gaussian noise source with zero mean and unity variance, whose variance is controlled by noise intensity $D$ as $\mathrm{Var}[\sqrt{2D}\cdot\xi_{n,i}(t)] = 2D$.
 We will denote $\mathrm{E}[\cdot]$ as the expected value and $\mathrm{Var}[\cdot]$ as variance of a random variable.
 The expected value of neuron's noisy output coincides with its noise free value $\mathrm{E}[y_{n,i}]=\mathrm{E}[x_{n,i}]$.
 The variance of signal with additive or multiplicative noise is $\mathrm{Var}[y_{n,i}]=2D_A+\mathrm{Var}[x_{n,i}]$ and $\mathrm{Var}[y_{n,i}]=2D_M\cdot\Big(\mathrm{E}^2[y_{n,i}] + \mathrm{Var}[x_{n,i}] \Big)$, respectively.
 Without noise-contamination in previous layers, both variances become $2D_A$ or $2D_M\cdot\mathrm{E}^2[y_{n,i}]$ \cite{Semenova2022}.

Furthermore, noise can be correlated or uncorrelated across numbers of neurons, such as all neurons in one layer.
 We use indices 'C' and 'U' to label these two features, see schematic illustration in Fig.~\ref{fig:FNN_scheme}(b).
 Combining all four noise types leads to the general description for the output of the $i$th neuron in layer $n$:
\begin{equation}
\begin{array}{c}
y_{n,i}(t) = \sqrt{2D^C_A}\xi^{C,A}_n(t) + \sqrt{2D^U_A}\xi^{U,A}_{n,i}(t) + \\
x_{n,i}(t)\cdot \Big(1+\sqrt{2D^C_M}\xi^{C,M}_n(t)\Big)\Big(1+\sqrt{2D^U_M}\xi^{U,M}_{n,i}(t)\Big).
\end{array}
\end{equation}

To characterize the noise level in numerical simulation, we use SNR, calculated as a ratio between expected value of the output signal and corresponding standard deviation or square root of variance \cite{Everitt1998}: $\mathrm{SNR}[\vec{y}^\mathrm{out}]=\mathrm{E}[\vec{y}^\mathrm{out}]/\sqrt{\mathrm{Var}[\vec{y}^\mathrm{out}]}$.
 In order to numerically determine the SNR, we repeat the same input signal $K=300$ times to calculate mean and standard deviation for each entry in the noise-less input sequence.

\section{Principles of network topology and noise accumulation}

\subsection{Linear vs. nonlinear FNNs}

\begin{figure}[h!]
\center{\includegraphics[width=\linewidth]{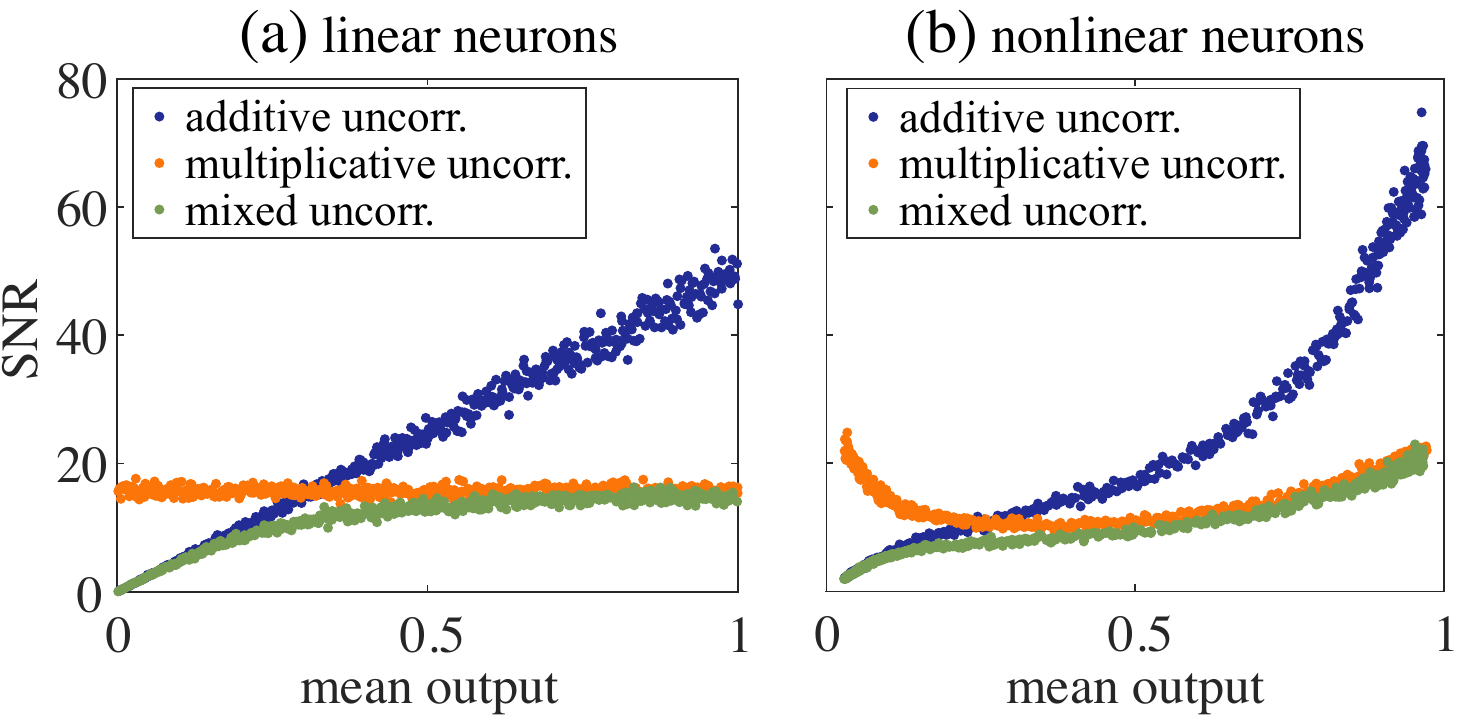}}
\caption[]{SNR in two FNNs which are schematically shown in Fig.~\ref{fig:FNN_scheme}(a) with one linear neurons in the input and output layers $I_1=I_3=1$. Neurons in hidden layer $I_2=100$ are linear in panel (a) and nonlinear with sigmoid activation function $f(x)=1/(1+\mathrm{exp}(-7(x-0.5)))$. Figures are prepared for additive (blue dots), multiplicative (orange) and mixed (green) noise with intensities $D^U_A=10^{-4}$, $D^U_M=10^{-3}$ }\label{fig:FNN_SNR}
\end{figure}

Nonlinearity can have a significant impact on noise propagation.
 In \cite{Semenova2019}, we showed that the FNN similar to Fig.~\ref{fig:FNN_scheme}(a) but with only linear neurons results in SNR curves as in Fig.~\ref{fig:FNN_SNR}(a) for additive (blue), multiplicative (orange) and mixed (green) uncorrelated noise.
 For FNNs with nonlinear neurons \cite{Semenova2022}, the SNR relationship intimately depends on particularities of the nonlinear activation functions, see Fig.~\ref{fig:FNN_SNR}(b) unsing the same color scheme.
 For both cases, the properties of mixed noise (additive \& multiplicative) is the superposition of both individual dependencies.
 The main overall result was that correlated noise accumulates stronger than uncorrelated noise.
 If, for example, connections are global and highly uniform, uncorrelated noise is essentially suppressed through averaging across the many connections.

\subsection{Input and output layers}

Highly relevant for a physical NN noise are its in and output layers \cite{Semenova2019,Semenova2022}.
 In particular for a single input neuron, i.e. scalar input information, all noise present at the input drives responses in the following layers, and can therefore not be suppressed through averaging.
 Similarly, noise-suppression through averaging along many network connections is impossible at the FNN's output, and noise in readout neurons is another major influence \cite{Semenova2019}.
 Placing relatively more resources to reduce hardware noise in the input and output layer is therefore an important guide of physical NN hardware design.
 However, such 'special' in and output neurons might not always be feasible or economic, or the attainable performance might be not sufficient for particular settings.
 We therefore propose several techniques that allow to further reduce noise accumulation without changing the properties of neurons themselves.

\subsection{Impact of intra-hidden layer connection topology}

In \cite{Semenova2022}, we considered trained FNNs and developed the analytical treatment enabling the accurate prediction of noise.
Importantly, our analytics show that accumulation of different noise types is greatly influenced by the connection matrices' statistics.
 Details of the analytical derivation can be found in Appendix \ref{sec:appendix}.

Noise propagation and accumulation is greatly influenced by the squared mean 
\begin{equation}\label{eq:SquaredMean}
\mu^2(\mathbf{W}^n)=\Big(\frac{1}{I_n I_{n+1}}\sum\limits_{i,j}W^n_{i,j}\Big)^2
\end{equation}
\noindent and the mean of the square
\begin{equation}\label{eq:MeanSquare}
\eta(\mathbf{W}^n)=\frac{1}{I_n I_{n+1}}\sum\limits_{i,j}(W^n_{i,j})^2    
\end{equation}
\noindent of connection matrix $\mathbf{W}^n$.
A hidden layer's noise-induced variance is determined by, both, noise in the current as well as by noise coming from previous layers.
 The impact of correlated noise in the current layer scales according to 
\begin{equation}\label{eq:CorrNoise}
I^2_n\cdot\mu^2(\mathbf{W}^{n}), 
\end{equation} 
\noindent while the impact of uncorrelated noise \emph{and} the noise from the previous layer scales according to 
\begin{equation}\label{eq:PreUnCorr}
 I_n \cdot\eta(\mathbf{W}^n),   
\end{equation}
\noindent see Ref.\cite{Semenova2022} and Appendix.
 There, by changing the statistics of $\mathbf{W}^n$, we can therefore greatly influence the accumulation of noise.

\begin{figure}[h!]
\center{\includegraphics[width=\linewidth]{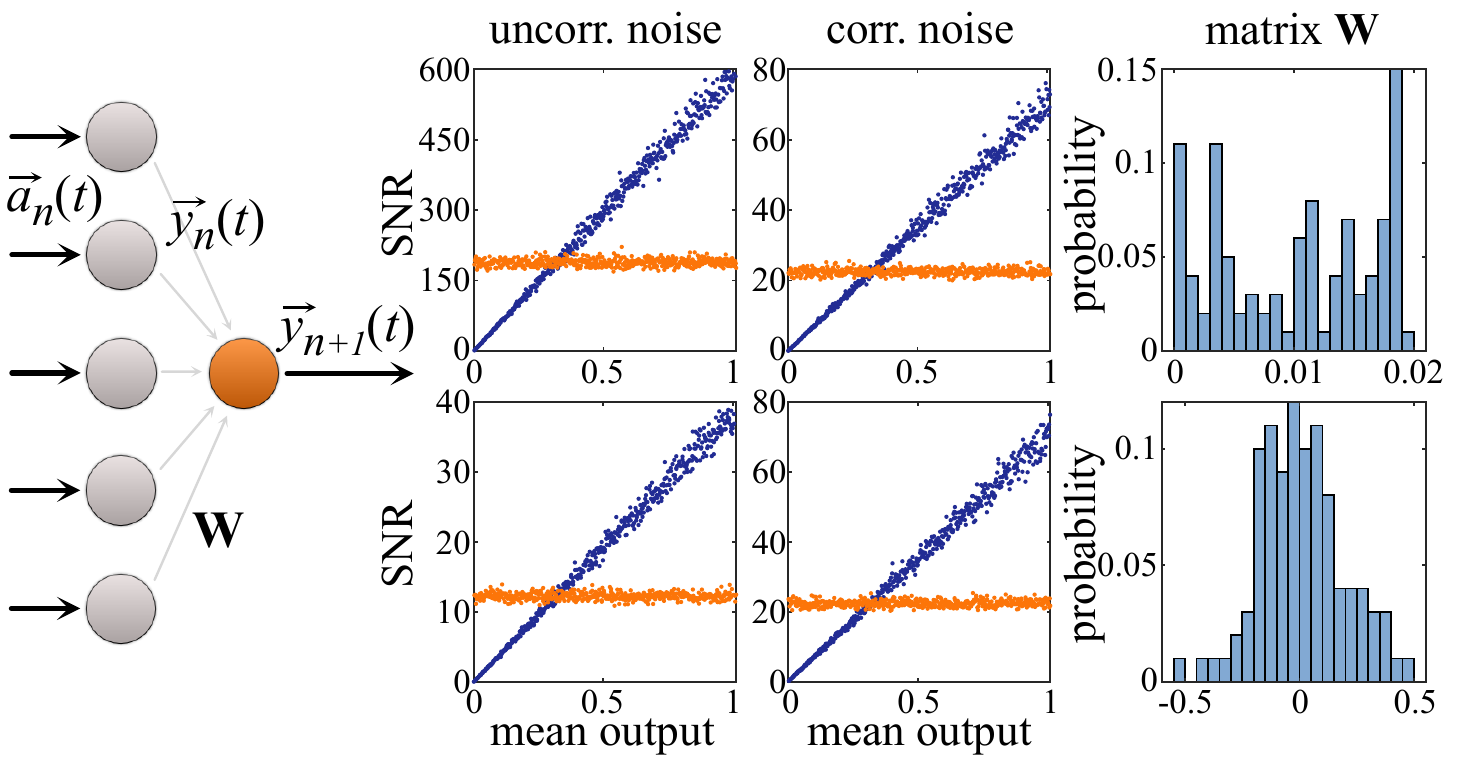}}
\caption[]{SNR for different noise intensities and connection matrices $\mathbf{W}$. Blue dots show the SNR curves with only additive noise, while orange dots are prepared for only multiplicative noise. The top panels correspond to the matrix with $I\mu^2(\mathbf{W})>\eta(\mathbf{W})$, namely $I\mu^2(\mathbf{W})=0.0103$, $\eta(\mathbf{W}) = 1.44\cdot 10^{-3}$. The bottom panel correspond to the opposite case when $I\mu^2(\mathbf{W})<\eta(\mathbf{W})$, namely $I\mu^2(\mathbf{W})=0.0101$, $\eta(\mathbf{W}) = 0.0340$. Noise intensities are $D^U_A=D^C_A=10^{-4}$, $D^U_M=D^C_M=10^{-3}$, $I=100$.}\label{fig:eta_mu}
\end{figure}

Figure~\ref{fig:eta_mu} shows the numerical results leveraging our findings.
 Here, we focus on the relevant aspects by only considering a FNN schematically illustrated in Fig.~\ref{fig:eta_mu}. 
 The layer consists of $I=100$ nonlinear neurons, and at each time iteration they receive the same input signal $u(t)$ randomly drawn from the interval [0;1].
 All neurons exhibit the same noisy additive and multiplicative noise that is in parts correlated as well as uncorrelated, parameters are given in the caption of Fig.~\ref{fig:eta_mu}.
 This noisy layer is connected to a single linear and noiseless output neuron according to connection matrix $\mathbf{W}$.

Figure \ref{fig:eta_mu} shows SNR curves for additive (blue) and multiplicative (orange) noise sources for two statistically different connection matrices.
 For a matrix for which $I \mu^2(\mathbf{W})> \eta(\mathbf{W})$ the accumulation of uncorrelated noise and noise from previous layers is effectively removed, see top panels in Fig.~\ref{fig:eta_mu}. 
 On the other hand, a matrix with $I \mu^2(\mathbf{W})< \eta(\mathbf{W})$ increases uncorrelated noise (bottom panels in Fig.~\ref{fig:eta_mu}), and the corresponding SNRs become lower.
 These relations between matrices do not influence correlated noise's contribution, and for comparable levels of correlated and uncorrelated noise, one will see mainly the impact of correlated noise for $I \mu^2(\mathbf{W})> \eta(\mathbf{W})$ and the one of uncorrelated noise if $I \mu^2(\mathbf{W})< \eta(\mathbf{W})$.
 An important conclusion is that if uncorrelated noise dominates, one can simply leverage learning (optimization) algorithms that force the system towards a topology with $I \mu^2(\mathbf{W})> \eta(\mathbf{W})$.
 A common mechanism for inducing correlating noise is a noisy power supply in a general sense.
 In electronics, this could be the circuit stabilising $V_{dd}$, while in optics this could be a pump or illumination source of photonic neurons.
 Since a general system will only have very few of such components, it appears feasible that these should receive an increased attention during the design stage.

\section{Ghost neurons for additive correlated noise mitigation}\label{sec:miti_ghost}

\begin{figure}[tbp] 
\center{\includegraphics[width=1\linewidth]{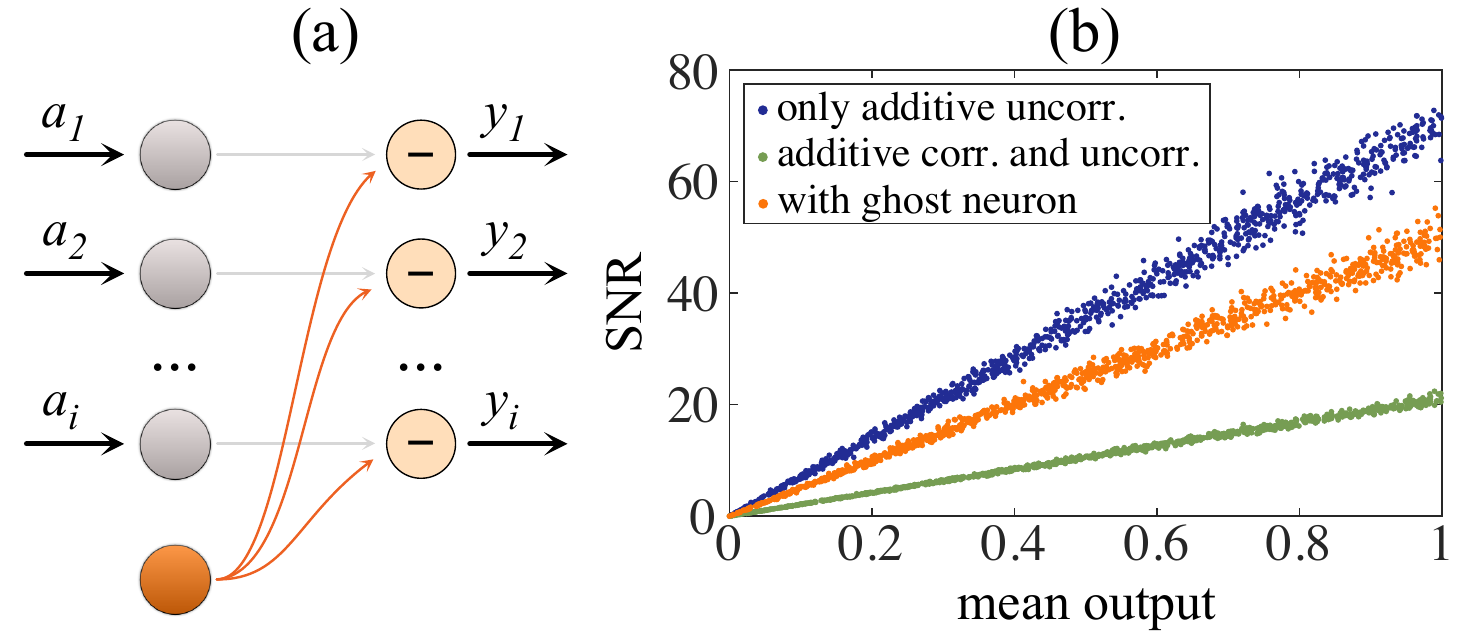}}
\caption{Schematic representation, how the ghost neuron can be added to the network with direct coupling (panel (a)). Panel (b) shows SNR for the case without ghost neuron with only additive uncorrelated noise (blue points), with both types of additive noise (green) and for the case with one ghost neuron (orange color). Noise intensities are $D^U_A=10^{-4}$, $D^C_A=10^{-3}$}\label{fig:miti_ghost_direct}
\end{figure}

Let us consider a FNN layer illustrated in Fig.~\ref{fig:miti_ghost_direct}(a) comprising of $I=100$ nonlinear and noisy neurons.
 Each neuron $i$ receives input signal $a_i$ emulating a neuron's input from the previous layer. Then the output of neuron $i$ including correlated and uncorrelated additive noise is
\begin{equation}\label{eq:ghost_out}
\begin{array}{c}
y_i=f(a_i)+\sqrt{2D^C_A}\xi^{C,A} +\sqrt{2D^U_A}\xi^{U,A}_i, \\
\mathrm{Var}[y_i]= 2D^C_A+2D^U_A .
\end{array}
\end{equation}
\noindent We now suppress additive noise and include an extra neuron with identical noise properties.
 Importantly, this \textit{ghost neuron} receives no input, but simply mimics the noise within the layer.
 The ghost neuron's output is then simply subtracted from each neuron's output, before this value $y_i$ propagates to the next later, which
 results in
\begin{equation}\label{eq:ghost_out2}
\begin{array}{c}
y_i= \Big(f(a_i)+\sqrt{2D^U_A}\xi^{U,A}_i -\sqrt{2D^U_A}\xi^{U,A}_g\Big),\\
\mathrm{Var}[y_i]= 4D^U_A .
\end{array}
\end{equation}

\begin{figure}[tbp] 
\center{\includegraphics[width=1\linewidth]{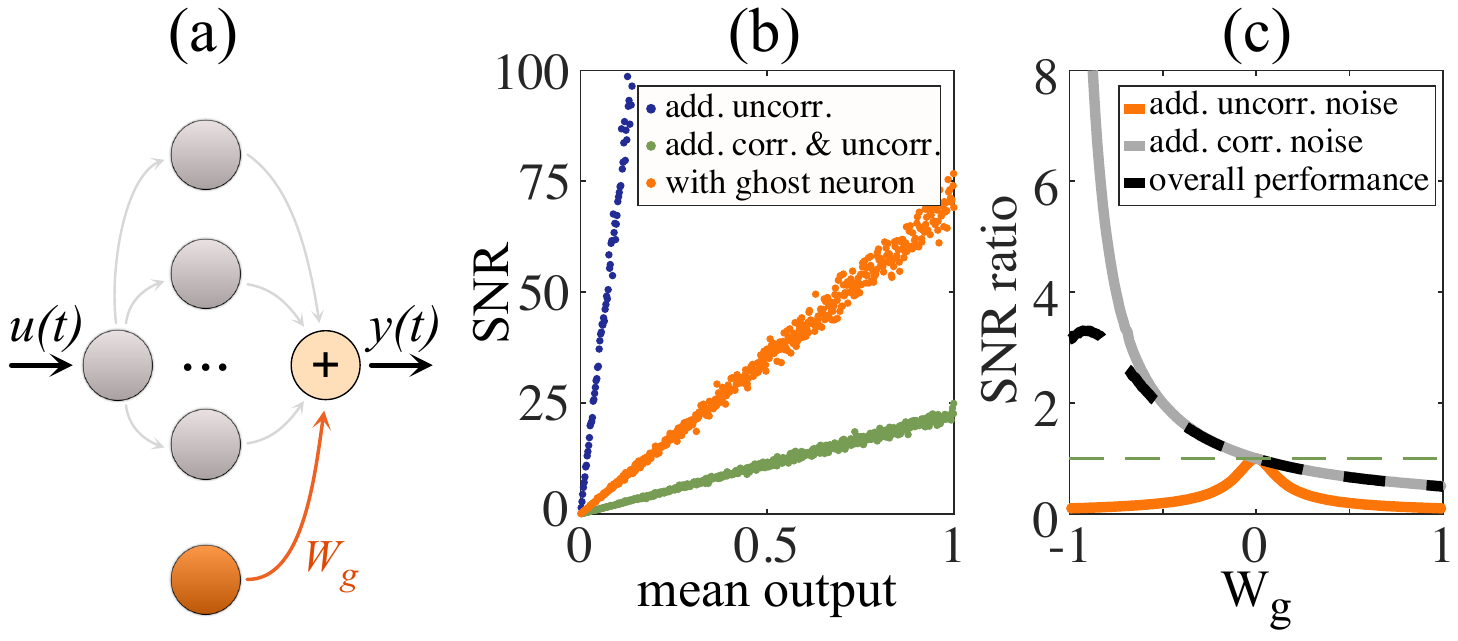}}
\caption{Noise mitigation with ghost neuron for the uniform coupling schematically shown in the panel (a) and the SNR obtained in numerical simulation for $W_g=-1$ (panel (b)). SNR is prepared for the case without ghost neuron with only additive uncorrelated noise (blue points), with both types of additive noise (green) and for the case with one ghost neuron (orange color). Panel (c) shows noise mitigation with ghost neuron for the uniform coupling shown depending on the weight of the ghost neuron $W_g$. Noise intensities are the same as in Fig.~\ref{fig:miti_ghost_direct}.}\label{fig:miti_ghost_uniform}
\end{figure}

As can be seen from Eq.~(\ref{eq:ghost_out2}), a ghost neuron fully suppresses correlated additive noise, yet the impact of uncorrelated additive noise is doubled.
 We confirm this in numerical simulation shown in Fig.~\ref{fig:miti_ghost_direct}(b). 
 However, as we showed before, uncorrelated noise can be suppressed leveraging coupling statistics, in particular $I\mu^2(\mathbf{W})>\eta(\mathbf{W})$.
 Rather than simply subtracting the ghost neuron's values as in Fig.~\ref{fig:miti_ghost_direct}(a), we now assign a weight to the ghost neuron's connection $W_g$, Fig.~\ref{fig:miti_ghost_uniform}(a).
 The output transforms into
\begin{equation}
\begin{array}{c}
y=\sum\limits^{I_n}_{j=1}W^n_{j1}\Big( f(u)+\sqrt{2D^U_A}\xi^{U,A}_{n,j}+\sqrt{2D^C_A}\xi^{C,A}_n  \Big)  + \\
 W_g\Big( \sqrt{2D^U_A}\xi^{U,A}_{g}+\sqrt{2D^C_A}\xi^{C,A}_n   \Big),
\end{array}
\end{equation}
and the corresponding variance is 
\begin{equation}\label{eq:miti_ghost_Wg_general}
\begin{array}{c}
\mathrm{Var}[y]=
\sum\limits^{I_n}_{j=1}(W^n_{j1})^2\cdot 2D^U_A + W^2_g\cdot 2D^U_A + \\ 
\Big(\sum\limits^{I_n}_{j=1}W^n_{j1}+W_g\Big)^2\cdot 2 D^C_A \approx \\
2D^U_A\cdot\Big(W^2_g + I_n\eta(\mathbf{W}^n)\Big) + 2D^C_A\cdot\Big(W_g + I_n\mu(\mathbf{W}^n)\Big)^2 .
\end{array}
\end{equation}
For the special case of a uniform connection matrix $W^n_{j1}=1/I_n$, the variance transforms to
\begin{equation}\label{eq:miti_ghost_Wg}
\mathrm{Var}[y] = 2D^U_A\cdot\bigg( \frac{1}{I_n}+W^2_g  \bigg) + 2D^C_A\cdot\Big(1+W_g\Big)^2.
\end{equation}
However, according to Eqs.~(\ref{eq:miti_ghost_Wg_general},\ref{eq:miti_ghost_Wg}), $W_g$ impacts correlated and uncorrelated noise differently.
 The multiplier of uncorrelated noise $\Big(W^2_g + I_n\eta(\mathbf{W}^n)\Big)$ shows that a ghost neuron increases the corresponding variance. 
 The multiplier of correlated noise $\Big(W_g + I_n\mu(\mathbf{W}^n)\Big)^2$ indicates that if $W_g=-I_n\mu(\mathbf{W}^n)$ or $W_g=-1$ for uniform connectivity, then correlated noise is fully suppressed. 
 Figure \ref{fig:miti_ghost_uniform}(b) numerically shows the case $W_g=-1$, which completely suppresses correlated additive noise, but at the same time increases uncorrelated noise.
 As a consequence, one needs to optimize $W_g$ in function of the different noise amplitudes.
 Figure \ref{fig:miti_ghost_uniform}(c) shows the averaged ratio between SNRs obtained with and without ghost neuron depending on its weight $W_g$. Three types of noise are considered: additive uncorrelated noise (orange), additive correlated noise (gray) and both noise types (black). The best overall performance can be achieved when $W_g=-1$.
 
\section{Pooling. Uncorrelated noise reduction}\label{sec:miti_pooling}

\begin{figure}[h!] 
\center{\includegraphics[width=1\linewidth]{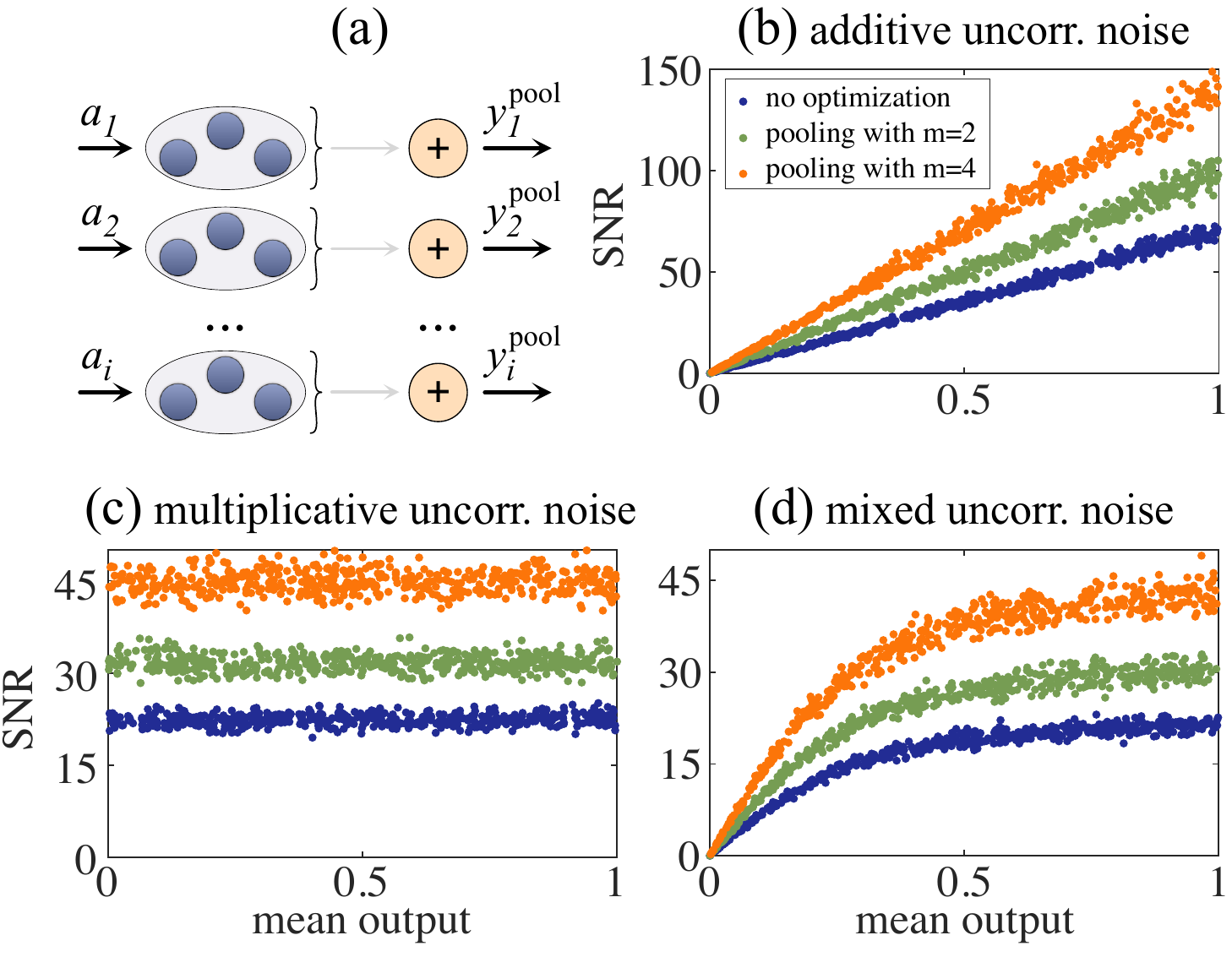}}
\caption{Scheme of noise mitigation with average pooling (a) and improvement of SNR due to pooling technique (b--d). Noise intensities are $D^U_M=10^{-3}$, $D^U_A=10^{-4}$.}\label{fig:miti_pool_SNR}
\end{figure}

In this section we discuss a common strategy to reduce uncorrelated noise without constraining connections $\mathbf{W}$.
 This method consists of combining several neurons into a distinct subgroups called pools.
 Each unit inside a pool of $m$ neurons receives the same input, see
 In Fig.~\ref{fig:miti_pool_SNR}(a). 
 The combined and hence averaged output signal of a pool is transmitted to the next layer.
 Each $k$th neuron of the $i$th group receiving the input signal $a_i$, has its own output value $y_{i,k}$ including noise and each group produces the averaged output $y^\mathrm{pool}_i=\frac{1}{m}\sum\limits^{m}_{k=1}y_{i,k}$. We used $m=3$ in Fig.~\ref{fig:miti_pool_SNR}(a).


For uncorrelated additive and multiplicative noise, the variance of the corresponding output without pooling is \cite{Semenova2019}
\begin{equation}\label{eq:miti_pool_var_no}
\mathrm{Var}[y_j]=2D^U_A+2D^U_M\cdot \mathrm{E}^2[y_j].
\end{equation}
 Using a pool with $m$ neurons then results in 
\begin{equation}\label{eq:miti_pool_var_with}
\begin{array}{c}
\mathrm{Var}[y^\mathrm{pool}_i]=\mathrm{Var}\bigg[  \frac{1}{m} \sum\limits^{m}_{k=1} y_{i,k}  \bigg] =
\frac{1}{m^2}\cdot \mathrm{Var}\bigg[\sum\limits^{m}_{k=1}y_{i,k}\bigg]= \\
\frac{1}{m^2}\cdot \sum\limits^{m}_{k=1}\Big(  2D^U_A + 2D^U_M\cdot \mathrm{E}^2[y_{i,k}]    \Big) =  \\
\frac{1}{m}\cdot \Big( 2D^U_A + 2D^U_M\cdot \mathrm{E}^2[y_{i}]    \Big),
\end{array}
\end{equation}
as the variance of the $i$th neuron pool output.
 Comparing Eqs.~(\ref{eq:miti_pool_var_no}, \ref{eq:miti_pool_var_with}), one can see that average pooling reduces the variance of uncorrelated additive and multiplicative noise  $m$ times, while the SNR improves by $\sqrt{m}$.
 Figure \ref{fig:miti_pool_SNR} shows the SNR for additive and multiplicative noise separately (panels (b) and (c), respectively) and for the mixed uncorrelated noise (d).

\section{Combining both techniques}\label{sec:miti_combo}

\begin{figure}[h!] 
\center{\includegraphics[width=1\linewidth]{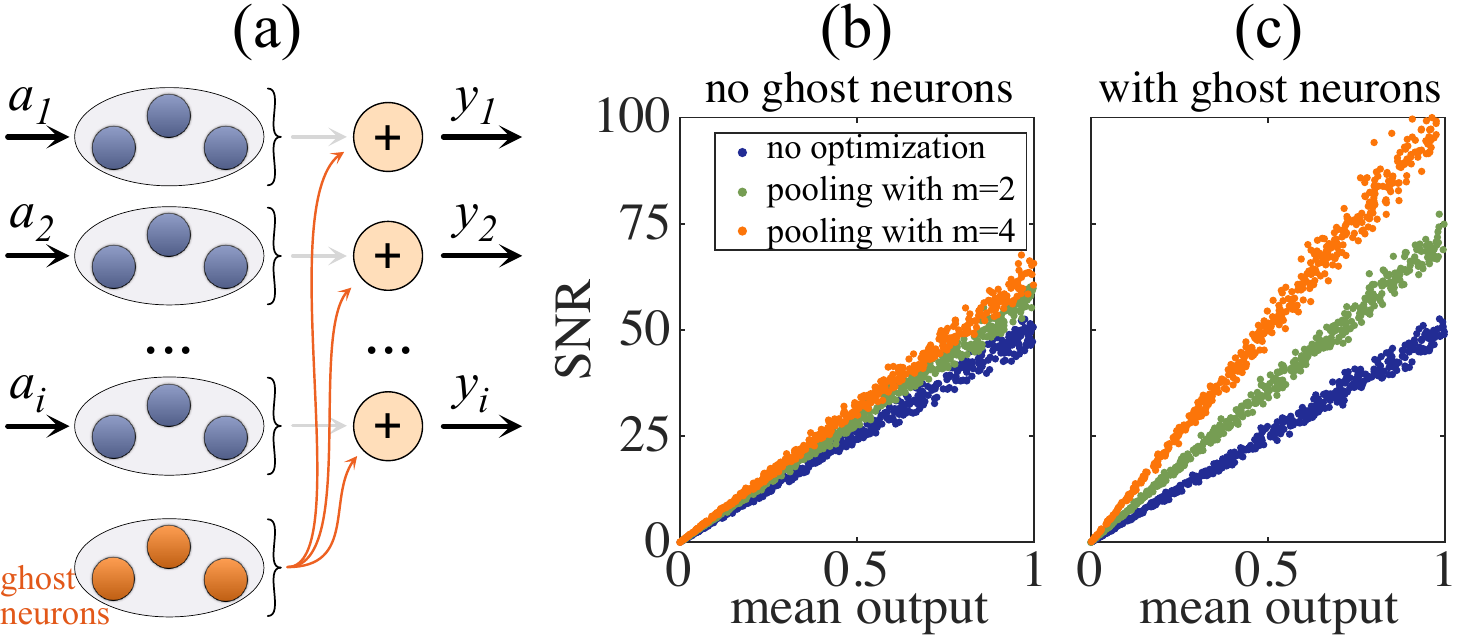}}
\caption{Scheme of noise mitigation with combined pooling ghost neuron technique (a) and improvement of SNR due to only pooling (b) and combined (c) techniques for additive correlated an uncorrelated noise. Noise intensities are $D^U_A=D^C_A=10^{-4}$.}\label{fig:miti_pool_ghost_SNR}
\end{figure}

Ghost neurons therefore remove correlated additive noise, while  uncorrelated noise can be addressed using average pooling.
 Crucially, both concepts can be combined, and Fig. \ref{fig:miti_pool_ghost_SNR}(a) illustrates the corresponding architecture, while panel (b) shows the SNR using average pooling in the case of, both, additive correlated and uncorrelated noise.
 Comparing Fig.~\ref{fig:miti_pool_SNR}(a) and Fig.~\ref{fig:miti_pool_ghost_SNR}(b), one can see the deteriorating effect of pooling when correlated noise is present.
 However, adding a ghost neuron substantially improves the situation, see Fig.~\ref{fig:miti_pool_ghost_SNR}(c).

\section{Application to trained network}\label{sec:trained}

In this section we apply the described above techniques to trained FNN.
 The noise-free network is trained to recognize MNIST handwritten digits from  \cite{LeCunSite} using the open-source python software library Keras \cite{Keras}, using a network consisting of three layers whose connections where optimized with standard error back propagation.
 The first layer receives the input image's $28\times 28$ pixels.
 The hidden layer has 100 nonlinear neurons with sigmoid activation function $f(x)=\frac{1}{1+e^{-x}}$, and the ten possible digits results in 10 nonlinear neurons with the same activation function in the hidden layer.
 The network's classification result is given by the output neuron with the largest value.
 With our proof-of-concept NN, we obtain a classification accuracy of 97.54\% for the test data without noise.

\begin{figure}[h!] 
\center{\includegraphics[width=1\linewidth]{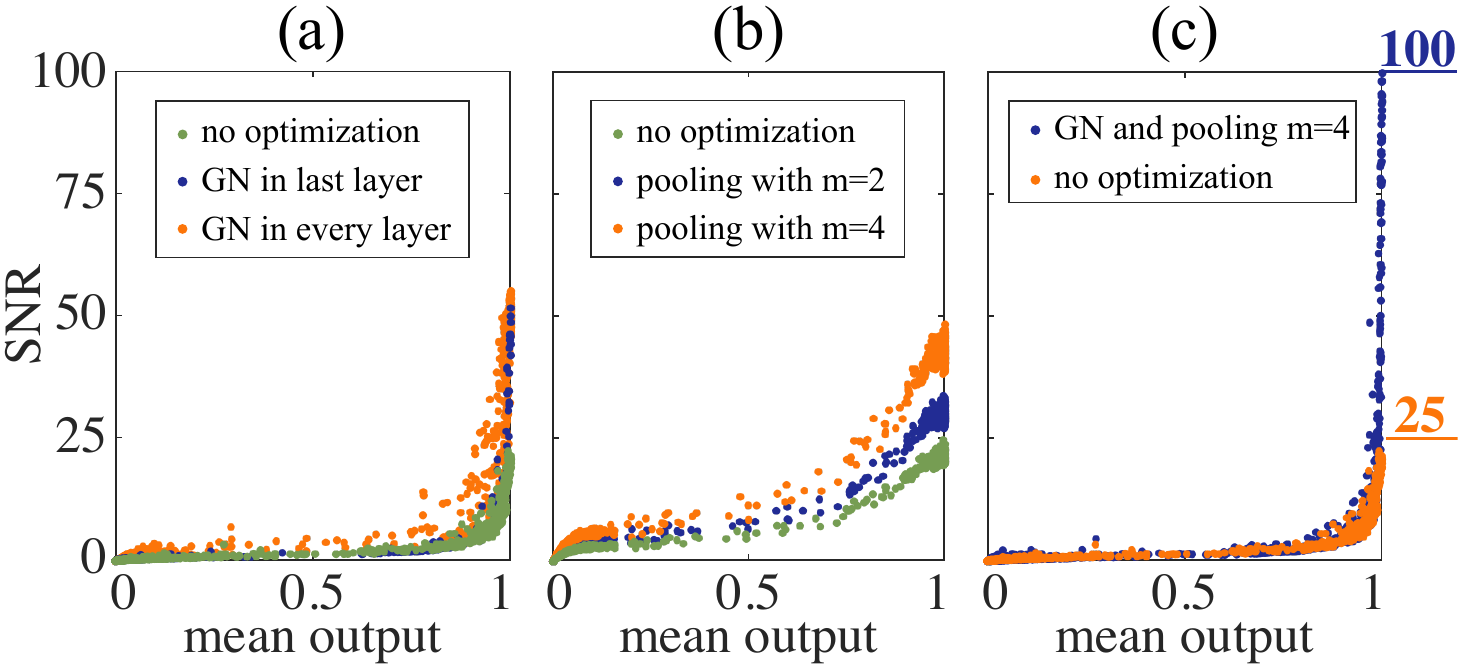}}
\caption{Noise reduction in FNN trained for MNIST digits recognition. Green dependencies in panels (a,b) are prepared without any noise reduction, and demonstrate the SNR of the output FNN signal. Panel (a) shows the noise reduction in trained network with additive uncorrelated $D^U_A=10^{-4}$ and correlated $D^C_A=10^{-3}$ noise using ghost neuron. Panel (b) demonstrates uncorrelated noise reduction with intensities $D^U_A=10^{-4}$ and $D^U_M=10^{-3}$ using pooling method with $m=2$ and $m=4$. Panel (c) shows the result combining both techniques for additive noise $D^U_A=10^{-4}$, $D^C_A=10^{-3}$.}\label{fig:trained_net}
\end{figure}

Figure~\ref{fig:trained_net}(a), green shows the SNR in the output layer for 500 randomly drawn digits without any noise mitigation strategy for $D^U_A=10^{-4}$ and $D^C_A=10^{-3}$.
 Figure~\ref{fig:trained_net}(a) shows the ghost neuron's impact when applied only in the final (blue data) as well as in all layers (orange data) with $W_g=-1$.
 Again, we  can see that mitigation of noise in the final layer is the most relevant.
 Secondly, we test pooling in a trained network with uncorrelated additive and multiplicative noise with noise intensities $D^U_A=10^{-4}$ and $D^U_M=10^{-3}$.
 The SNR without ($m=1$, green data) and with average pooling ($m=2$ for blue data and $m=4$ for orange data) is shown in Fig.~\ref{fig:trained_net}(b).
 However, we found almost no difference between pooling in all layers or only in the final one, which is because the strong suppression of uncorrelated noise by a densely connected network consequence of training, for which $I\mu^2(\mathbf{W})>\eta(\mathbf{W})$.
 We numerically confirmed that the SNR in our trained network is improved by a factor $\sqrt{m}$ for $m=2$ and $m=4$.

Finally, Fig.~\ref{fig:trained_net}(c) shows the SNR for combination of both techniques of ghost neuron in the last layer and pooling with $m=4$ for FNN with additive noise $D^U_A=10^{-4}$, $D^C_A=10^{-3}$. 
 Panel (c) demonstrates SNR with combined optimization (blue) and without it (orange), providing maximum SNR values 100 and 25, respectively. 
 Thus, combining technique leads to a 4-fold SNR improvement and consequently a 16-fold variance reduction.
 
All previous conclusions regarding the improvement of the noisy FNN were made with respect to SNR. 
 However, the accuracy is more important characteristics for classification and recognition tasks. For the noise-free FNN it is 97.54\%, while it drops to 92.97\% for noisy FNN with additive noise $D^U_A=10^{-4}$, $D^C_A=10^{-3}$ . 
 Using the combined technique form the previous paragraph, the accuracy can be improved slightly to 93.1\%.
 Meanwhile, the best performance can be achieved when using adaptive ghost neuron weights depending on matrices statistics: $W^n_{gi}=-\sum\limits^{I_n}_{j=1} W^n_{ij}$. If these ghost neurons are added to every layer optimized with pooling, then the range of SNR values remains the same as in Fig.~\ref{fig:trained_net}(c), but the accuracy becomes 97.49\%, which much closer to the noise-free FNN.

\section{Conclusions}

We have proposed several noise reduction strategies specifically leveraging our previous analytical insights obtained in \cite{Semenova2019,Semenova2022}, mitigating uncorrelated noise and additive correlated noise.
 First, we show how the the particular statistics of connection matrices allow the mitigation of particular noise types.
 Such strategies can be used to amend optimization (learning) algorithms.
 We go beyond and introduce two complementary techniques of the case when statistics of intra-layer connections cannot be modified. 
 Correlated additive noise can be removed using ghost neurons, while average pooling works well for, both, uncorrelated additive and multiplicative noise without impacting correlated noise.
 Furthermore, we show how both techniques can be combined to form a comprehensive topology to suppress noise on a physical NN's hardware level.
 All above techniques were successfully applied to a NN for MNIST handwritten digit recognition, where they showed a reduction in the noise level in agreement to our analytical descriptions and almost complete noise suppression in terms of network accuracy.

\begin{acknowledgments}
N. Semenova is supported by Russian Science Foundation (Project No. 21-72-00002). \end{acknowledgments}

\section*{Data Availability Statement}
The data that support the findings of this study are available from the corresponding author upon reasonable request.

\appendix*
\section{Importance of connection matrices statistics}\label{sec:appendix}
In order to illustrate the accumulation of noise, let us consider the vector of signals coming from noisy layer $n$ to $(n+1)$:
\begin{equation}\label{eq:app:z}
\vec{a}_{n+1}=\mathbf{W}^n\cdot \vec{y}_n, \text{ \ \ \ \ or \ \ \ \ } a_{n+1,i}=\sum\limits^{I_n}_{j=1} W^n_{ij}\cdot y_{n,j}.
\end{equation}
According to nomenclature of the main part of article, this value further transforms to $\vec{x}_{n+1}=f(\vec{a}_{n+1})$ after activation function and finally to $\vec{y}_{n+1}$ after the noise impact.

Substituting the noise to $y_{n,j}$, Eq.~(\ref{eq:app:z}) transforms to
\begin{equation}\label{eq:app:z_noise}
\begin{array}{c}
a_{n+1,i}=\sum\limits^{I_n}_{j=1} W^n_{ij}\cdot \Big(\sqrt{2D^C_A}\xi^{C,A}_n+\sqrt{2D^U_A}\xi^{U,A}_{n,j} \Big) + \\ 
\sum\limits^{I_n}_{j=1} W^n_{ij}x_{n,j}\cdot \Big(1+\sqrt{2D^C_M}\xi^{C,M}_n\Big) \Big(1+\sqrt{2D^U_M}\xi^{U,M}_{n,j} \Big).
\end{array}
\end{equation}
All terms and multipliers of correlated noise do not depend on index $j$ and they can be therefore moved out of sums:
\begin{equation}\label{eq:app:z_noise2}
\begin{array}{c}
a_{n+1,i}=\sqrt{2D^C_A}\xi^{C,A}_n\cdot\sum\limits^{I_n}_{j=1} W^n_{ij} + \sqrt{2D^U_A}\cdot\sum\limits^{I_n}_{j=1} W^n_{ij}\xi^{U,A}_{n,j}\\
+\big(1+\sqrt{2D^C_M}\xi^{C,M}_n\big) \cdot\sum\limits^{I_n}_{j=1} W^n_{ij}x_{n,j}\Big(1+\sqrt{2D^U_M}\xi^{U,M}_{n,j} \Big).
\end{array}
\end{equation}
The variance of this noisy signal will be determined based on the basic arithmetic principles of calculating the variance of random variables \cite{Montgomery2002} such as:
\begin{equation}\nonumber
\begin{array}{c}
\mathrm{Var}[c\cdot \xi]=c^2\cdot\mathrm{Var}[\xi]; \ \ \ \mathrm{Var}[\xi+c]=\mathrm{Var}[\xi]; \\ \mathrm{Var}[\xi\pm\zeta]=\mathrm{Var}[\xi]+\mathrm{Var}[\zeta]; \\
\mathrm{Var}[\xi\cdot\zeta]=\big(\mathrm{E}^2[\xi]+\mathrm{Var}[\xi]\big)\mathrm{Var}[\zeta] + \mathrm{E}[\zeta]\mathrm{Var}[\xi], 
\end{array}
\end{equation}
where $\xi$ and $\zeta$ are some uncorrelated random variables, $c$ is some constant or noise-free variable. Then the variance of Eq.~(\ref{eq:app:z_noise2}) is
\begin{equation}\nonumber
\begin{array}{c}
\mathrm{Var}[a_{n+1,i}] = 2D^C_A\Big( \sum\limits^{I_n}_{j=1} W^n_{ij} \Big)^2 + 2D^U_A \sum\limits^{I_n}_{j=1} \big( W^n_{ij} \big)^2 + \\
\big(1+2D^C_M\big)\cdot \mathrm{Var}\Big[ \sum\limits^{I_n}_{j=1} W^n_{ij}x_{n,j}\big(1+ \sqrt{2D^U_M}\xi^{U,M}_n \big) \Big]+ \\
2D^C_M\cdot \mathrm{E}^2\Big[ \sum\limits^{I_n}_{j=1} W^n_{ij}x_{n,j}\big(1+ \sqrt{2D^U_M}\xi^{U,M}_n \big) \Big] = \\

2D^C_A\Big( \sum\limits^{I_n}_{j=1} W^n_{ij} \Big)^2 + 2D^C_M\cdot\Big( \sum\limits^{I_n}_{j=1} W^n_{ij}\mathrm{E}[x_{n,j}] \Big)^2 + \\
2D^U_A \sum\limits^{I_n}_{j=1} \big( W^n_{ij} \big)^2  +  
2D^U_M\big(1+2D^C_M\big) \sum\limits^{I_n}_{j=1} \big(W^n_{ij}\big)^2\mathrm{E}^2[x_{n,j}] \\
+ \big(1+2D^C_M\big)\big(1+2D^U_M\big)\cdot \sum\limits^{I_n}_{j=1} \big(W^n_{ij}\big)^2\mathrm{Var}[x_{n,j}].
\end{array}
\end{equation}
For simplification, we assume that $\sum\limits^{I_n}_{j=1}\big(W^n_{ij}\big)^2 \approx I_n\cdot \eta(\mathbf{W}^n)$ and $\Big(\sum\limits^{I_n}_{j=1}W^n_{ij}\Big)^2 \approx I^2_n\cdot \mu^2(\mathbf{W}^n)$, where $\eta(\cdot)$ is the mean of the square and $\mu(\cdot)$ is the mean (see Eqs.~(\ref{eq:SquaredMean},\ref{eq:MeanSquare}), main text). Then
\begin{equation}\label{eq:app:final_var}
\begin{array}{c}
\mathrm{Var}[a_{n+1,i}]\approx 2D^C_A\cdot I^2_n\mu^2\big(\mathbf{W}^n\big) + 2D^U_A\cdot I_n\eta\big(\mathbf{W}^n\big) +\\
 2D^C_M \mu^2\big(\mathrm{E}[\vec{x}_n]\big)\cdot I^2_n\mu^2\big(\mathbf{W}^n\big) + \\
 2D^U_M(1+2D^C_M)\eta\big(\mathrm{E}[\vec{x}_n]\big)\cdot I_n\eta\big(\mathbf{W}^n\big)+ \\
(1+2D^C_M)(1+2D^U_M)\eta(\mathbf{W}^n)\cdot \mathrm{Var}[\vec{x}_n].
\end{array}
\end{equation}
We will not go into detail about the last term of Eq.~(\ref{eq:app:final_var}) as it is not the subject of this article, and it has been described and analyzed in Ref.\cite{Semenova2022}. It is clearly seen, that all rest terms with $I^2_n\mu^2\big(\mathbf{W}^n\big)$ are related to correlated noise as:
\begin{equation}\label{eq:app:corr_noise}
I^2_n\mu^2\big(\mathbf{W}^n\big)\cdot\Big\{ 2D^C_A+2D^C_M\cdot\mu^2\big( \mathrm{E}[\vec{x}_n] \big)\Big\},
\end{equation}
while terms with $I_n\eta\big(\mathbf{W}^n\big)$ are
\begin{equation}\label{eq:app:uncorr_noise}
I_n\eta\big(\mathbf{W}^n\big)\cdot\Big\{ 2D^U_A+2D^U_M\big(1+2D^C_M\big)\cdot\eta\big( \mathrm{E}[\vec{x}_n] \big)\Big\}.
\end{equation}
Comparing Eqs.~(\ref{eq:app:corr_noise}) and (\ref{eq:app:uncorr_noise}) one can see that if $I_n\mu^2\big(\mathbf{W}^n\big)>\eta\big(\mathbf{W}^n\big)$, then the impact of uncorrelated noise is less than the correlated noise when noise intensities are the same $D^U_A=D^C_A$, $D^U_M=D^C_M$.


\begin{thebibliography}{38}%
\makeatletter
\providecommand \@ifxundefined [1]{%
 \@ifx{#1\undefined}
}%
\providecommand \@ifnum [1]{%
 \ifnum #1\expandafter \@firstoftwo
 \else \expandafter \@secondoftwo
 \fi
}%
\providecommand \@ifx [1]{%
 \ifx #1\expandafter \@firstoftwo
 \else \expandafter \@secondoftwo
 \fi
}%
\providecommand \natexlab [1]{#1}%
\providecommand \enquote  [1]{``#1''}%
\providecommand \bibnamefont  [1]{#1}%
\providecommand \bibfnamefont [1]{#1}%
\providecommand \citenamefont [1]{#1}%
\providecommand \href@noop [0]{\@secondoftwo}%
\providecommand \href [0]{\begingroup \@sanitize@url \@href}%
\providecommand \@href[1]{\@@startlink{#1}\@@href}%
\providecommand \@@href[1]{\endgroup#1\@@endlink}%
\providecommand \@sanitize@url [0]{\catcode `\\12\catcode `\$12\catcode
  `\&12\catcode `\#12\catcode `\^12\catcode `\_12\catcode `\%12\relax}%
\providecommand \@@startlink[1]{}%
\providecommand \@@endlink[0]{}%
\providecommand \url  [0]{\begingroup\@sanitize@url \@url }%
\providecommand \@url [1]{\endgroup\@href {#1}{\urlprefix }}%
\providecommand \urlprefix  [0]{URL }%
\providecommand \Eprint [0]{\href }%
\providecommand \doibase [0]{http://dx.doi.org/}%
\providecommand \selectlanguage [0]{\@gobble}%
\providecommand \bibinfo  [0]{\@secondoftwo}%
\providecommand \bibfield  [0]{\@secondoftwo}%
\providecommand \translation [1]{[#1]}%
\providecommand \BibitemOpen [0]{}%
\providecommand \bibitemStop [0]{}%
\providecommand \bibitemNoStop [0]{.\EOS\space}%
\providecommand \EOS [0]{\spacefactor3000\relax}%
\providecommand \BibitemShut  [1]{\csname bibitem#1\endcsname}%
\let\auto@bib@innerbib\@empty
\bibitem [{\citenamefont {LeCun}, \citenamefont {Bengio},\ and\ \citenamefont
  {Hinton}(2015)}]{Lecun2015}%
  \BibitemOpen
  \bibfield  {author} {\bibinfo {author} {\bibfnamefont {Y.}~\bibnamefont
  {LeCun}}, \bibinfo {author} {\bibfnamefont {Y.}~\bibnamefont {Bengio}}, \
  and\ \bibinfo {author} {\bibfnamefont {G.}~\bibnamefont {Hinton}},\
  }\bibfield  {title} {\enquote {\bibinfo {title} {Deep learning},}\ }\href
  {\doibase 10.1038/nature14539} {\bibfield  {journal} {\bibinfo  {journal}
  {Nature}\ }\textbf {\bibinfo {volume} {521}},\ \bibinfo {pages} {436--444}
  (\bibinfo {year} {2015})}\BibitemShut {NoStop}%
\bibitem [{\citenamefont {Krizhevsky}, \citenamefont {Sutskever},\ and\
  \citenamefont {Hinton}(2017)}]{Krizhevsky2017}%
  \BibitemOpen
  \bibfield  {author} {\bibinfo {author} {\bibfnamefont {A.}~\bibnamefont
  {Krizhevsky}}, \bibinfo {author} {\bibfnamefont {I.}~\bibnamefont
  {Sutskever}}, \ and\ \bibinfo {author} {\bibfnamefont {G.~E.}\ \bibnamefont
  {Hinton}},\ }\bibfield  {title} {\enquote {\bibinfo {title} {Imagenet
  classification with deep convolutional neural networks},}\ }\href {\doibase
  10.1145/3065386} {\bibfield  {journal} {\bibinfo  {journal} {Commun. ACM}\
  }\textbf {\bibinfo {volume} {60}},\ \bibinfo {pages} {84--90} (\bibinfo
  {year} {2017})}\BibitemShut {NoStop}%
\bibitem [{\citenamefont {Maturana}\ and\ \citenamefont
  {Scherer}(2015)}]{Maturana2015}%
  \BibitemOpen
  \bibfield  {author} {\bibinfo {author} {\bibfnamefont {D.}~\bibnamefont
  {Maturana}}\ and\ \bibinfo {author} {\bibfnamefont {S.}~\bibnamefont
  {Scherer}},\ }\bibfield  {title} {\enquote {\bibinfo {title} {Voxnet: A 3d
  convolutional neural network for real-time object recognition},}\ }in\ \href
  {\doibase 10.1109/IROS.2015.7353481} {\emph {\bibinfo {booktitle} {2015
  IEEE/RSJ International Conference on Intelligent Robots and Systems
  (IROS)}}}\ (\bibinfo {year} {2015})\ pp.\ \bibinfo {pages}
  {922--928}\BibitemShut {NoStop}%
\bibitem [{\citenamefont {Graves}, \citenamefont {Mohamed},\ and\ \citenamefont
  {Hinton}(2013)}]{Graves2013}%
  \BibitemOpen
  \bibfield  {author} {\bibinfo {author} {\bibfnamefont {A.}~\bibnamefont
  {Graves}}, \bibinfo {author} {\bibfnamefont {A.-r.}\ \bibnamefont {Mohamed}},
  \ and\ \bibinfo {author} {\bibfnamefont {G.}~\bibnamefont {Hinton}},\
  }\bibfield  {title} {\enquote {\bibinfo {title} {Speech recognition with deep
  recurrent neural networks},}\ }in\ \href {\doibase
  10.1109/ICASSP.2013.6638947} {\emph {\bibinfo {booktitle} {2013 IEEE
  International Conference on Acoustics, Speech and Signal Processing}}}\
  (\bibinfo {year} {2013})\ pp.\ \bibinfo {pages} {6645--6649}\BibitemShut
  {NoStop}%
\bibitem [{\citenamefont {Kar}\ and\ \citenamefont {Moura}(2009)}]{Kar2009}%
  \BibitemOpen
  \bibfield  {author} {\bibinfo {author} {\bibfnamefont {S.}~\bibnamefont
  {Kar}}\ and\ \bibinfo {author} {\bibfnamefont {J.~M.~F.}\ \bibnamefont
  {Moura}},\ }\bibfield  {title} {\enquote {\bibinfo {title} {Distributed
  consensus algorithms in sensor networks with imperfect communication: Link
  failures and channel noise},}\ }\href {\doibase 10.1109/TSP.2008.2007111}
  {\bibfield  {journal} {\bibinfo  {journal} {IEEE Transactions on Signal
  Processing}\ }\textbf {\bibinfo {volume} {57}},\ \bibinfo {pages} {355--369}
  (\bibinfo {year} {2009})}\BibitemShut {NoStop}%
\bibitem [{\citenamefont {Gupta}\ \emph {et~al.}(2015)\citenamefont {Gupta},
  \citenamefont {Agrawal}, \citenamefont {Gopalakrishnan},\ and\ \citenamefont
  {Narayanan}}]{Gupta2015}%
  \BibitemOpen
  \bibfield  {author} {\bibinfo {author} {\bibfnamefont {S.}~\bibnamefont
  {Gupta}}, \bibinfo {author} {\bibfnamefont {A.}~\bibnamefont {Agrawal}},
  \bibinfo {author} {\bibfnamefont {K.}~\bibnamefont {Gopalakrishnan}}, \ and\
  \bibinfo {author} {\bibfnamefont {P.}~\bibnamefont {Narayanan}},\ }\bibfield
  {title} {\enquote {\bibinfo {title} {{Deep Learning with Limited Numerical
  Precision}},}\ }\href {\doibase 10.1109/72.80206} {\bibfield  {journal}
  {\bibinfo  {journal} {Proceedings of the 32nd International Conference on
  International Conference on Machine Learning}\ }\textbf {\bibinfo {volume}
  {37}},\ \bibinfo {pages} {1737--1746} (\bibinfo {year} {2015})}\BibitemShut
  {NoStop}%
\bibitem [{\citenamefont {Wright}\ \emph {et~al.}(2022)\citenamefont {Wright},
  \citenamefont {Onodera}, \citenamefont {Stein}, \citenamefont {Wang},
  \citenamefont {Schachter}, \citenamefont {Hu},\ and\ \citenamefont
  {McMahon}}]{Wright2022}%
  \BibitemOpen
  \bibfield  {author} {\bibinfo {author} {\bibfnamefont {L.~G.}\ \bibnamefont
  {Wright}}, \bibinfo {author} {\bibfnamefont {T.}~\bibnamefont {Onodera}},
  \bibinfo {author} {\bibfnamefont {M.~M.}\ \bibnamefont {Stein}}, \bibinfo
  {author} {\bibfnamefont {T.}~\bibnamefont {Wang}}, \bibinfo {author}
  {\bibfnamefont {D.~T.}\ \bibnamefont {Schachter}}, \bibinfo {author}
  {\bibfnamefont {Z.}~\bibnamefont {Hu}}, \ and\ \bibinfo {author}
  {\bibfnamefont {P.~L.}\ \bibnamefont {McMahon}},\ }\bibfield  {title}
  {\enquote {\bibinfo {title} {{Deep physical neural networks trained with
  backpropagation}},}\ }\href {\doibase 10.1038/s41586-021-04223-6} {\bibfield
  {journal} {\bibinfo  {journal} {Nature}\ }\textbf {\bibinfo {volume} {601}},\
  \bibinfo {pages} {549--555} (\bibinfo {year} {2022})}\BibitemShut {NoStop}%
\bibitem [{\citenamefont {Markovi{\'{c}}}\ \emph {et~al.}(2020)\citenamefont
  {Markovi{\'{c}}}, \citenamefont {Mizrahi}, \citenamefont {Querlioz},\ and\
  \citenamefont {Grollier}}]{Markovic2020}%
  \BibitemOpen
  \bibfield  {author} {\bibinfo {author} {\bibfnamefont {D.}~\bibnamefont
  {Markovi{\'{c}}}}, \bibinfo {author} {\bibfnamefont {A.}~\bibnamefont
  {Mizrahi}}, \bibinfo {author} {\bibfnamefont {D.}~\bibnamefont {Querlioz}}, \
  and\ \bibinfo {author} {\bibfnamefont {J.}~\bibnamefont {Grollier}},\
  }\bibfield  {title} {\enquote {\bibinfo {title} {{Physics for neuromorphic
  computing}},}\ }\href {\doibase 10.1038/s42254-020-0208-2} {\bibfield
  {journal} {\bibinfo  {journal} {Nature Reviews Physics}\ }\textbf {\bibinfo
  {volume} {2}},\ \bibinfo {pages} {499--510} (\bibinfo {year}
  {2020})}\BibitemShut {NoStop}%
\bibitem [{\citenamefont {Boahen}(2017)}]{Boahen2017}%
  \BibitemOpen
  \bibfield  {author} {\bibinfo {author} {\bibfnamefont {K.}~\bibnamefont
  {Boahen}},\ }\bibfield  {title} {\enquote {\bibinfo {title} {{A neuromorph's
  Prospectus}},}\ }\href {\doibase 10.1109/MCSE.2017.33} {\bibfield  {journal}
  {\bibinfo  {journal} {Computing in Science \& Engineering}\ }\textbf
  {\bibinfo {volume} {19}},\ \bibinfo {pages} {14--28} (\bibinfo {year}
  {2017})}\BibitemShut {NoStop}%
\bibitem [{\citenamefont {Wang}\ \emph {et~al.}(2018)\citenamefont {Wang},
  \citenamefont {Joshi}, \citenamefont {Savel'Ev}, \citenamefont {Song},
  \citenamefont {Midya}, \citenamefont {Li}, \citenamefont {Rao}, \citenamefont
  {Yan}, \citenamefont {Asapu}, \citenamefont {Zhuo}, \citenamefont {Jiang},
  \citenamefont {Lin}, \citenamefont {Li}, \citenamefont {Yoon}, \citenamefont
  {Upadhyay}, \citenamefont {Zhang}, \citenamefont {Hu}, \citenamefont
  {Strachan}, \citenamefont {Barnell}, \citenamefont {Wu}, \citenamefont {Wu},
  \citenamefont {Williams}, \citenamefont {Xia},\ and\ \citenamefont
  {Yang}}]{Wang2018}%
  \BibitemOpen
  \bibfield  {author} {\bibinfo {author} {\bibfnamefont {Z.}~\bibnamefont
  {Wang}}, \bibinfo {author} {\bibfnamefont {S.}~\bibnamefont {Joshi}},
  \bibinfo {author} {\bibfnamefont {S.}~\bibnamefont {Savel'Ev}}, \bibinfo
  {author} {\bibfnamefont {W.}~\bibnamefont {Song}}, \bibinfo {author}
  {\bibfnamefont {R.}~\bibnamefont {Midya}}, \bibinfo {author} {\bibfnamefont
  {Y.}~\bibnamefont {Li}}, \bibinfo {author} {\bibfnamefont {M.}~\bibnamefont
  {Rao}}, \bibinfo {author} {\bibfnamefont {P.}~\bibnamefont {Yan}}, \bibinfo
  {author} {\bibfnamefont {S.}~\bibnamefont {Asapu}}, \bibinfo {author}
  {\bibfnamefont {Y.}~\bibnamefont {Zhuo}}, \bibinfo {author} {\bibfnamefont
  {H.}~\bibnamefont {Jiang}}, \bibinfo {author} {\bibfnamefont
  {P.}~\bibnamefont {Lin}}, \bibinfo {author} {\bibfnamefont {C.}~\bibnamefont
  {Li}}, \bibinfo {author} {\bibfnamefont {J.~H.}\ \bibnamefont {Yoon}},
  \bibinfo {author} {\bibfnamefont {N.~K.}\ \bibnamefont {Upadhyay}}, \bibinfo
  {author} {\bibfnamefont {J.}~\bibnamefont {Zhang}}, \bibinfo {author}
  {\bibfnamefont {M.}~\bibnamefont {Hu}}, \bibinfo {author} {\bibfnamefont
  {J.~P.}\ \bibnamefont {Strachan}}, \bibinfo {author} {\bibfnamefont
  {M.}~\bibnamefont {Barnell}}, \bibinfo {author} {\bibfnamefont
  {Q.}~\bibnamefont {Wu}}, \bibinfo {author} {\bibfnamefont {H.}~\bibnamefont
  {Wu}}, \bibinfo {author} {\bibfnamefont {R.~S.}\ \bibnamefont {Williams}},
  \bibinfo {author} {\bibfnamefont {Q.}~\bibnamefont {Xia}}, \ and\ \bibinfo
  {author} {\bibfnamefont {J.~J.}\ \bibnamefont {Yang}},\ }\bibfield  {title}
  {\enquote {\bibinfo {title} {{Fully memristive neural networks for pattern
  classification with unsupervised learning}},}\ }\href {\doibase
  10.1038/s41928-018-0023-2} {\bibfield  {journal} {\bibinfo  {journal} {Nature
  Electronics}\ }\textbf {\bibinfo {volume} {1}},\ \bibinfo {pages} {137--145}
  (\bibinfo {year} {2018})}\BibitemShut {NoStop}%
\bibitem [{\citenamefont {Lin}\ \emph {et~al.}(2020)\citenamefont {Lin},
  \citenamefont {Li}, \citenamefont {Wang}, \citenamefont {Li}, \citenamefont
  {Jiang}, \citenamefont {Song}, \citenamefont {Rao}, \citenamefont {Zhuo},
  \citenamefont {Upadhyay}, \citenamefont {Barnell}, \citenamefont {Wu},
  \citenamefont {Yang},\ and\ \citenamefont {Xia}}]{Lin2020}%
  \BibitemOpen
  \bibfield  {author} {\bibinfo {author} {\bibfnamefont {P.}~\bibnamefont
  {Lin}}, \bibinfo {author} {\bibfnamefont {C.}~\bibnamefont {Li}}, \bibinfo
  {author} {\bibfnamefont {Z.}~\bibnamefont {Wang}}, \bibinfo {author}
  {\bibfnamefont {Y.}~\bibnamefont {Li}}, \bibinfo {author} {\bibfnamefont
  {H.}~\bibnamefont {Jiang}}, \bibinfo {author} {\bibfnamefont
  {W.}~\bibnamefont {Song}}, \bibinfo {author} {\bibfnamefont {M.}~\bibnamefont
  {Rao}}, \bibinfo {author} {\bibfnamefont {Y.}~\bibnamefont {Zhuo}}, \bibinfo
  {author} {\bibfnamefont {N.~K.}\ \bibnamefont {Upadhyay}}, \bibinfo {author}
  {\bibfnamefont {M.}~\bibnamefont {Barnell}}, \bibinfo {author} {\bibfnamefont
  {Q.}~\bibnamefont {Wu}}, \bibinfo {author} {\bibfnamefont {J.~J.}\
  \bibnamefont {Yang}}, \ and\ \bibinfo {author} {\bibfnamefont
  {Q.}~\bibnamefont {Xia}},\ }\bibfield  {title} {\enquote {\bibinfo {title}
  {{Three-dimensional memristor circuits as complex neural networks}},}\ }\href
  {\doibase 10.1038/s41928-020-0397-9} {\bibfield  {journal} {\bibinfo
  {journal} {Nature Electronics}\ }\textbf {\bibinfo {volume} {3}},\ \bibinfo
  {pages} {225--232} (\bibinfo {year} {2020})}\BibitemShut {NoStop}%
\bibitem [{\citenamefont {Xia}\ and\ \citenamefont {Yang}(2019)}]{Xia2019}%
  \BibitemOpen
  \bibfield  {author} {\bibinfo {author} {\bibfnamefont {Q.}~\bibnamefont
  {Xia}}\ and\ \bibinfo {author} {\bibfnamefont {J.~J.}\ \bibnamefont {Yang}},\
  }\bibfield  {title} {\enquote {\bibinfo {title} {{Memristive crossbar arrays
  for brain-inspired computing}},}\ }\href {\doibase 10.1038/s41563-019-0291-x}
  {\bibfield  {journal} {\bibinfo  {journal} {Nature Materials}\ }\textbf
  {\bibinfo {volume} {18}},\ \bibinfo {pages} {309--323} (\bibinfo {year}
  {2019})}\BibitemShut {NoStop}%
\bibitem [{\citenamefont {Feldmann}\ \emph {et~al.}(2021)\citenamefont
  {Feldmann}, \citenamefont {Youngblood}, \citenamefont {Karpov}, \citenamefont
  {Gehring}, \citenamefont {Li}, \citenamefont {Stappers}, \citenamefont
  {Le~Gallo}, \citenamefont {Fu}, \citenamefont {Lukashchuk}, \citenamefont
  {Raja}, \citenamefont {Liu}, \citenamefont {Wright}, \citenamefont
  {Sebastian}, \citenamefont {Kippenberg}, \citenamefont {Pernice},\ and\
  \citenamefont {Bhaskaran}}]{Feldmann2021}%
  \BibitemOpen
  \bibfield  {author} {\bibinfo {author} {\bibfnamefont {J.}~\bibnamefont
  {Feldmann}}, \bibinfo {author} {\bibfnamefont {N.}~\bibnamefont
  {Youngblood}}, \bibinfo {author} {\bibfnamefont {M.}~\bibnamefont {Karpov}},
  \bibinfo {author} {\bibfnamefont {H.}~\bibnamefont {Gehring}}, \bibinfo
  {author} {\bibfnamefont {X.}~\bibnamefont {Li}}, \bibinfo {author}
  {\bibfnamefont {M.}~\bibnamefont {Stappers}}, \bibinfo {author}
  {\bibfnamefont {M.}~\bibnamefont {Le~Gallo}}, \bibinfo {author}
  {\bibfnamefont {X.}~\bibnamefont {Fu}}, \bibinfo {author} {\bibfnamefont
  {A.}~\bibnamefont {Lukashchuk}}, \bibinfo {author} {\bibfnamefont {A.~S.}\
  \bibnamefont {Raja}}, \bibinfo {author} {\bibfnamefont {J.}~\bibnamefont
  {Liu}}, \bibinfo {author} {\bibfnamefont {C.~D.}\ \bibnamefont {Wright}},
  \bibinfo {author} {\bibfnamefont {A.}~\bibnamefont {Sebastian}}, \bibinfo
  {author} {\bibfnamefont {T.~J.}\ \bibnamefont {Kippenberg}}, \bibinfo
  {author} {\bibfnamefont {W.~H.~P.}\ \bibnamefont {Pernice}}, \ and\ \bibinfo
  {author} {\bibfnamefont {H.}~\bibnamefont {Bhaskaran}},\ }\bibfield  {title}
  {\enquote {\bibinfo {title} {Parallel convolutional processing using an
  integrated photonic tensor core},}\ }\href {\doibase
  10.1038/s41586-020-03070-1} {\bibfield  {journal} {\bibinfo  {journal}
  {Nature}\ }\textbf {\bibinfo {volume} {589}},\ \bibinfo {pages} {52--58}
  (\bibinfo {year} {2021})}\BibitemShut {NoStop}%
\bibitem [{\citenamefont {Brunner}\ \emph {et~al.}(2013)\citenamefont
  {Brunner}, \citenamefont {Soriano}, \citenamefont {Mirasso},\ and\
  \citenamefont {Fischer}}]{Brunner2013a}%
  \BibitemOpen
  \bibfield  {author} {\bibinfo {author} {\bibfnamefont {D.}~\bibnamefont
  {Brunner}}, \bibinfo {author} {\bibfnamefont {M.~C.}\ \bibnamefont
  {Soriano}}, \bibinfo {author} {\bibfnamefont {C.~R.}\ \bibnamefont
  {Mirasso}}, \ and\ \bibinfo {author} {\bibfnamefont {I.}~\bibnamefont
  {Fischer}},\ }\bibfield  {title} {\enquote {\bibinfo {title} {{Parallel
  photonic information processing at gigabyte per second data rates using
  transient states}},}\ }\href@noop {} {\bibfield  {journal} {\bibinfo
  {journal} {Nature communications}\ }\textbf {\bibinfo {volume} {4}},\
  \bibinfo {pages} {1364} (\bibinfo {year} {2013})}\BibitemShut {NoStop}%
\bibitem [{\citenamefont {Nguimdo}\ \emph {et~al.}(2020)\citenamefont
  {Nguimdo}, \citenamefont {Antonik}, \citenamefont {Marsal},\ and\
  \citenamefont {Rontani}}]{Nguimdo2020}%
  \BibitemOpen
  \bibfield  {author} {\bibinfo {author} {\bibfnamefont {R.~M.}\ \bibnamefont
  {Nguimdo}}, \bibinfo {author} {\bibfnamefont {P.}~\bibnamefont {Antonik}},
  \bibinfo {author} {\bibfnamefont {N.}~\bibnamefont {Marsal}}, \ and\ \bibinfo
  {author} {\bibfnamefont {D.}~\bibnamefont {Rontani}},\ }\bibfield  {title}
  {\enquote {\bibinfo {title} {Impact of optical coherence on the performance
  of large-scale spatiotemporal photonic reservoir computing systems},}\ }\href
  {\doibase 10.1364/OE.400546} {\bibfield  {journal} {\bibinfo  {journal} {Opt.
  Express}\ }\textbf {\bibinfo {volume} {28}},\ \bibinfo {pages} {27989--28005}
  (\bibinfo {year} {2020})}\BibitemShut {NoStop}%
\bibitem [{\citenamefont {Huang}\ \emph {et~al.}(2022)\citenamefont {Huang},
  \citenamefont {Sorger}, \citenamefont {Miscuglio}, \citenamefont {Al-Qadasi},
  \citenamefont {Mukherjee}, \citenamefont {Lampe}, \citenamefont {Nichols},
  \citenamefont {Tait}, \citenamefont {de~Lima}, \citenamefont {Marquez},
  \citenamefont {Wang}, \citenamefont {Chrostowski}, \citenamefont {Fok},
  \citenamefont {Brunner}, \citenamefont {Fan}, \citenamefont {Shekhar},
  \citenamefont {Prucnal},\ and\ \citenamefont {Shastri}}]{Huang2022}%
  \BibitemOpen
  \bibfield  {author} {\bibinfo {author} {\bibfnamefont {C.}~\bibnamefont
  {Huang}}, \bibinfo {author} {\bibfnamefont {V.~J.}\ \bibnamefont {Sorger}},
  \bibinfo {author} {\bibfnamefont {M.}~\bibnamefont {Miscuglio}}, \bibinfo
  {author} {\bibfnamefont {M.}~\bibnamefont {Al-Qadasi}}, \bibinfo {author}
  {\bibfnamefont {A.}~\bibnamefont {Mukherjee}}, \bibinfo {author}
  {\bibfnamefont {L.}~\bibnamefont {Lampe}}, \bibinfo {author} {\bibfnamefont
  {M.}~\bibnamefont {Nichols}}, \bibinfo {author} {\bibfnamefont {A.~N.}\
  \bibnamefont {Tait}}, \bibinfo {author} {\bibfnamefont {T.~F.}\ \bibnamefont
  {de~Lima}}, \bibinfo {author} {\bibfnamefont {B.~A.}\ \bibnamefont
  {Marquez}}, \bibinfo {author} {\bibfnamefont {J.}~\bibnamefont {Wang}},
  \bibinfo {author} {\bibfnamefont {L.}~\bibnamefont {Chrostowski}}, \bibinfo
  {author} {\bibfnamefont {M.~P.}\ \bibnamefont {Fok}}, \bibinfo {author}
  {\bibfnamefont {D.}~\bibnamefont {Brunner}}, \bibinfo {author} {\bibfnamefont
  {S.}~\bibnamefont {Fan}}, \bibinfo {author} {\bibfnamefont {S.}~\bibnamefont
  {Shekhar}}, \bibinfo {author} {\bibfnamefont {P.~R.}\ \bibnamefont
  {Prucnal}}, \ and\ \bibinfo {author} {\bibfnamefont {B.~J.}\ \bibnamefont
  {Shastri}},\ }\bibfield  {title} {\enquote {\bibinfo {title} {Prospects and
  applications of photonic neural networks},}\ }\href {\doibase
  10.1080/23746149.2021.1981155} {\bibfield  {journal} {\bibinfo  {journal}
  {Advances in Physics: X}\ }\textbf {\bibinfo {volume} {7}},\ \bibinfo {pages}
  {1981155} (\bibinfo {year} {2022})},\ \Eprint
  {http://arxiv.org/abs/https://doi.org/10.1080/23746149.2021.1981155}
  {https://doi.org/10.1080/23746149.2021.1981155} \BibitemShut {NoStop}%
\bibitem [{\citenamefont {Wang}\ \emph {et~al.}(2022)\citenamefont {Wang},
  \citenamefont {Ma}, \citenamefont {Wright}, \citenamefont {Onodera},
  \citenamefont {Richard},\ and\ \citenamefont {McMahon}}]{Wang2022}%
  \BibitemOpen
  \bibfield  {author} {\bibinfo {author} {\bibfnamefont {T.}~\bibnamefont
  {Wang}}, \bibinfo {author} {\bibfnamefont {S.-Y.}\ \bibnamefont {Ma}},
  \bibinfo {author} {\bibfnamefont {L.~G.}\ \bibnamefont {Wright}}, \bibinfo
  {author} {\bibfnamefont {T.}~\bibnamefont {Onodera}}, \bibinfo {author}
  {\bibfnamefont {B.~C.}\ \bibnamefont {Richard}}, \ and\ \bibinfo {author}
  {\bibfnamefont {P.~L.}\ \bibnamefont {McMahon}},\ }\bibfield  {title}
  {\enquote {\bibinfo {title} {An optical neural network using less than 1
  photon per multiplication},}\ }\href {\doibase 10.1038/s41467-021-27774-8}
  {\bibfield  {journal} {\bibinfo  {journal} {Nature Communications}\ }\textbf
  {\bibinfo {volume} {13}},\ \bibinfo {pages} {123} (\bibinfo {year}
  {2022})}\BibitemShut {NoStop}%
\bibitem [{\citenamefont {Panda}\ and\ \citenamefont
  {Hegde}(2022)}]{Panda2022}%
  \BibitemOpen
  \bibfield  {author} {\bibinfo {author} {\bibfnamefont {S.~S.}\ \bibnamefont
  {Panda}}\ and\ \bibinfo {author} {\bibfnamefont {R.~S.}\ \bibnamefont
  {Hegde}},\ }\bibfield  {title} {\enquote {\bibinfo {title} {Fault tolerance
  and noise immunity in freespace diffractive optical neural networks},}\
  }\href {\doibase 10.1088/2631-8695/ac4832} {\bibfield  {journal} {\bibinfo
  {journal} {Engineering Research Express}\ }\textbf {\bibinfo {volume} {4}},\
  \bibinfo {pages} {011301} (\bibinfo {year} {2022})}\BibitemShut {NoStop}%
\bibitem [{Tor(2017)}]{Torrejon2017}%
  \BibitemOpen
  \bibfield  {title} {\enquote {\bibinfo {title} {{Neuromorphic computing with
  nanoscale spintronic oscillators}},}\ }\href@noop {} {\bibfield  {journal}
  {\bibinfo  {journal} {Nature}\ }\textbf {\bibinfo {volume} {547}},\ \bibinfo
  {pages} {428--431} (\bibinfo {year} {2017})}\BibitemShut {NoStop}%
\bibitem [{\citenamefont {Psaltis}\ \emph {et~al.}(1990)\citenamefont
  {Psaltis}, \citenamefont {Brady}, \citenamefont {Gu},\ and\ \citenamefont
  {Lin}}]{Psaltis1990}%
  \BibitemOpen
  \bibfield  {author} {\bibinfo {author} {\bibfnamefont {D.}~\bibnamefont
  {Psaltis}}, \bibinfo {author} {\bibfnamefont {D.}~\bibnamefont {Brady}},
  \bibinfo {author} {\bibfnamefont {X.-G.}\ \bibnamefont {Gu}}, \ and\ \bibinfo
  {author} {\bibfnamefont {S.}~\bibnamefont {Lin}},\ }\bibfield  {title}
  {\enquote {\bibinfo {title} {{Holography in artificial neural networks}},}\
  }\href {\doibase 10.1038/343325a0} {\bibfield  {journal} {\bibinfo  {journal}
  {Nature}\ }\textbf {\bibinfo {volume} {343}},\ \bibinfo {pages} {325--330}
  (\bibinfo {year} {1990})}\BibitemShut {NoStop}%
\bibitem [{\citenamefont {Bueno}\ \emph {et~al.}(2018)\citenamefont {Bueno},
  \citenamefont {Maktoobi}, \citenamefont {Froehly}, \citenamefont {Fischer},
  \citenamefont {Jacquot}, \citenamefont {Larger},\ and\ \citenamefont
  {Brunner}}]{Bueno2018}%
  \BibitemOpen
  \bibfield  {author} {\bibinfo {author} {\bibfnamefont {J.}~\bibnamefont
  {Bueno}}, \bibinfo {author} {\bibfnamefont {S.}~\bibnamefont {Maktoobi}},
  \bibinfo {author} {\bibfnamefont {L.}~\bibnamefont {Froehly}}, \bibinfo
  {author} {\bibfnamefont {I.}~\bibnamefont {Fischer}}, \bibinfo {author}
  {\bibfnamefont {M.}~\bibnamefont {Jacquot}}, \bibinfo {author} {\bibfnamefont
  {L.}~\bibnamefont {Larger}}, \ and\ \bibinfo {author} {\bibfnamefont
  {D.}~\bibnamefont {Brunner}},\ }\bibfield  {title} {\enquote {\bibinfo
  {title} {{Reinforcement Learning in a large scale photonic Recurrent Neural
  Network}},}\ }\href@noop {} {\bibfield  {journal} {\bibinfo  {journal}
  {Optica}\ }\textbf {\bibinfo {volume} {5}},\ \bibinfo {pages} {756 -- 760}
  (\bibinfo {year} {2018})}\BibitemShut {NoStop}%
\bibitem [{\citenamefont {Lin}\ \emph {et~al.}(2018)\citenamefont {Lin},
  \citenamefont {Rivenson}, \citenamefont {Yardimci}, \citenamefont {Veli},
  \citenamefont {Jarrahi},\ and\ \citenamefont {Ozcan}}]{Lin2018}%
  \BibitemOpen
  \bibfield  {author} {\bibinfo {author} {\bibfnamefont {X.}~\bibnamefont
  {Lin}}, \bibinfo {author} {\bibfnamefont {Y.}~\bibnamefont {Rivenson}},
  \bibinfo {author} {\bibfnamefont {N.~T.}\ \bibnamefont {Yardimci}}, \bibinfo
  {author} {\bibfnamefont {M.}~\bibnamefont {Veli}}, \bibinfo {author}
  {\bibfnamefont {M.}~\bibnamefont {Jarrahi}}, \ and\ \bibinfo {author}
  {\bibfnamefont {A.}~\bibnamefont {Ozcan}},\ }\bibfield  {title} {\enquote
  {\bibinfo {title} {{All-Optical Machine Learning Using Diffractive Deep
  Neural Networks}},}\ }\href {\doibase 10.1126/science.aat8084} {\bibfield
  {journal} {\bibinfo  {journal} {Science}\ }\textbf {\bibinfo {volume} {26}},\
  \bibinfo {pages} {1--20} (\bibinfo {year} {2018})}\BibitemShut {NoStop}%
\bibitem [{\citenamefont {Shen}\ \emph {et~al.}(2017)\citenamefont {Shen},
  \citenamefont {Harris}, \citenamefont {Skirlo}, \citenamefont {Prabhu},
  \citenamefont {Baehr-Jones}, \citenamefont {Hochberg}, \citenamefont {Sun},
  \citenamefont {Zhao}, \citenamefont {Larochelle}, \citenamefont {Englund},\
  and\ \citenamefont {Soljacic}}]{Shen2017}%
  \BibitemOpen
  \bibfield  {author} {\bibinfo {author} {\bibfnamefont {Y.}~\bibnamefont
  {Shen}}, \bibinfo {author} {\bibfnamefont {N.~C.}\ \bibnamefont {Harris}},
  \bibinfo {author} {\bibfnamefont {S.}~\bibnamefont {Skirlo}}, \bibinfo
  {author} {\bibfnamefont {M.}~\bibnamefont {Prabhu}}, \bibinfo {author}
  {\bibfnamefont {T.}~\bibnamefont {Baehr-Jones}}, \bibinfo {author}
  {\bibfnamefont {M.}~\bibnamefont {Hochberg}}, \bibinfo {author}
  {\bibfnamefont {X.}~\bibnamefont {Sun}}, \bibinfo {author} {\bibfnamefont
  {S.}~\bibnamefont {Zhao}}, \bibinfo {author} {\bibfnamefont {H.}~\bibnamefont
  {Larochelle}}, \bibinfo {author} {\bibfnamefont {D.}~\bibnamefont {Englund}},
  \ and\ \bibinfo {author} {\bibfnamefont {M.}~\bibnamefont {Soljacic}},\
  }\bibfield  {title} {\enquote {\bibinfo {title} {{Deep Learning with Coherent
  Nanophotonic Circuits}},}\ }\href@noop {} {\bibfield  {journal} {\bibinfo
  {journal} {Nature Photonics}\ }\textbf {\bibinfo {volume} {11}},\ \bibinfo
  {pages} {441--446} (\bibinfo {year} {2017})}\BibitemShut {NoStop}%
\bibitem [{\citenamefont {Tait}\ \emph {et~al.}(2017)\citenamefont {Tait},
  \citenamefont {{De Lima}}, \citenamefont {Zhou}, \citenamefont {Wu},
  \citenamefont {Nahmias}, \citenamefont {Shastri},\ and\ \citenamefont
  {Prucnal}}]{Tait2017}%
  \BibitemOpen
  \bibfield  {author} {\bibinfo {author} {\bibfnamefont {A.~N.}\ \bibnamefont
  {Tait}}, \bibinfo {author} {\bibfnamefont {T.~F.}\ \bibnamefont {{De Lima}}},
  \bibinfo {author} {\bibfnamefont {E.}~\bibnamefont {Zhou}}, \bibinfo {author}
  {\bibfnamefont {A.~X.}\ \bibnamefont {Wu}}, \bibinfo {author} {\bibfnamefont
  {M.~A.}\ \bibnamefont {Nahmias}}, \bibinfo {author} {\bibfnamefont {B.~J.}\
  \bibnamefont {Shastri}}, \ and\ \bibinfo {author} {\bibfnamefont {P.~R.}\
  \bibnamefont {Prucnal}},\ }\bibfield  {title} {\enquote {\bibinfo {title}
  {{Neuromorphic photonic networks using silicon photonic weight banks}},}\
  }\href@noop {} {\bibfield  {journal} {\bibinfo  {journal} {Scientific
  Reports}\ }\textbf {\bibinfo {volume} {7}},\ \bibinfo {pages} {1--10}
  (\bibinfo {year} {2017})}\BibitemShut {NoStop}%
\bibitem [{\citenamefont {Moughames}\ \emph
  {et~al.}(2020{\natexlab{a}})\citenamefont {Moughames}, \citenamefont {Porte},
  \citenamefont {Thiel}, \citenamefont {Ulliac}, \citenamefont {Larger},
  \citenamefont {Jacquot}, \citenamefont {Kadic},\ and\ \citenamefont
  {Brunner}}]{Moughames2020}%
  \BibitemOpen
  \bibfield  {author} {\bibinfo {author} {\bibfnamefont {J.}~\bibnamefont
  {Moughames}}, \bibinfo {author} {\bibfnamefont {X.}~\bibnamefont {Porte}},
  \bibinfo {author} {\bibfnamefont {M.}~\bibnamefont {Thiel}}, \bibinfo
  {author} {\bibfnamefont {G.}~\bibnamefont {Ulliac}}, \bibinfo {author}
  {\bibfnamefont {L.}~\bibnamefont {Larger}}, \bibinfo {author} {\bibfnamefont
  {M.}~\bibnamefont {Jacquot}}, \bibinfo {author} {\bibfnamefont
  {M.}~\bibnamefont {Kadic}}, \ and\ \bibinfo {author} {\bibfnamefont
  {D.}~\bibnamefont {Brunner}},\ }\bibfield  {title} {\enquote {\bibinfo
  {title} {Three-dimensional waveguide interconnects for scalable integration
  of photonic neural networks},}\ }\href {\doibase 10.1364/OPTICA.388205}
  {\bibfield  {journal} {\bibinfo  {journal} {Optica}\ }\textbf {\bibinfo
  {volume} {7}},\ \bibinfo {pages} {640--646} (\bibinfo {year}
  {2020}{\natexlab{a}})}\BibitemShut {NoStop}%
\bibitem [{\citenamefont {{Dinc, Niyazi Ulas}}, \citenamefont {{Psaltis,
  Demetri}},\ and\ \citenamefont {{Brunner, Daniel}}(2020)}]{Dinc2020}%
  \BibitemOpen
  \bibfield  {author} {\bibinfo {author} {\bibnamefont {{Dinc, Niyazi Ulas}}},
  \bibinfo {author} {\bibnamefont {{Psaltis, Demetri}}}, \ and\ \bibinfo
  {author} {\bibnamefont {{Brunner, Daniel}}},\ }\bibfield  {title} {\enquote
  {\bibinfo {title} {Optical neural networks: The 3d connection},}\ }\href
  {\doibase 10.1051/photon/202010434} {\bibfield  {journal} {\bibinfo
  {journal} {Photoniques}\ ,\ \bibinfo {pages} {34--38}} (\bibinfo {year}
  {2020})}\BibitemShut {NoStop}%
\bibitem [{\citenamefont {Moughames}\ \emph
  {et~al.}(2020{\natexlab{b}})\citenamefont {Moughames}, \citenamefont {Porte},
  \citenamefont {Larger}, \citenamefont {Jacquot}, \citenamefont {Kadic},\ and\
  \citenamefont {Brunner}}]{Moughames2020a}%
  \BibitemOpen
  \bibfield  {author} {\bibinfo {author} {\bibfnamefont {J.}~\bibnamefont
  {Moughames}}, \bibinfo {author} {\bibfnamefont {X.}~\bibnamefont {Porte}},
  \bibinfo {author} {\bibfnamefont {L.}~\bibnamefont {Larger}}, \bibinfo
  {author} {\bibfnamefont {M.}~\bibnamefont {Jacquot}}, \bibinfo {author}
  {\bibfnamefont {M.}~\bibnamefont {Kadic}}, \ and\ \bibinfo {author}
  {\bibfnamefont {D.}~\bibnamefont {Brunner}},\ }\bibfield  {title} {\enquote
  {\bibinfo {title} {3d printed multimode-splitters for photonic
  interconnects},}\ }\href {\doibase 10.1364/OME.402974} {\bibfield  {journal}
  {\bibinfo  {journal} {Opt. Mater. Express}\ }\textbf {\bibinfo {volume}
  {10}},\ \bibinfo {pages} {2952--2961} (\bibinfo {year}
  {2020}{\natexlab{b}})}\BibitemShut {NoStop}%
\bibitem [{\citenamefont {Dolenko}\ and\ \citenamefont
  {Card}(1993)}]{Dolenko1993}%
  \BibitemOpen
  \bibfield  {author} {\bibinfo {author} {\bibfnamefont {B.}~\bibnamefont
  {Dolenko}}\ and\ \bibinfo {author} {\bibfnamefont {H.}~\bibnamefont {Card}},\
  }\bibfield  {title} {\enquote {\bibinfo {title} {Neural learning in analogue
  hardware: effects of component variation from fabrication and from noise},}\
  }\href@noop {} {\bibfield  {journal} {\bibinfo  {journal} {Electronics
  letters}\ }\textbf {\bibinfo {volume} {29}},\ \bibinfo {pages} {693--694}
  (\bibinfo {year} {1993})}\BibitemShut {NoStop}%
\bibitem [{\citenamefont {Misra}\ and\ \citenamefont {Saha}(2010)}]{Misra2010}%
  \BibitemOpen
  \bibfield  {author} {\bibinfo {author} {\bibfnamefont {J.}~\bibnamefont
  {Misra}}\ and\ \bibinfo {author} {\bibfnamefont {I.}~\bibnamefont {Saha}},\
  }\bibfield  {title} {\enquote {\bibinfo {title} {Artificial neural networks
  in hardware: A survey of two decades of progress},}\ }\href {\doibase
  https://doi.org/10.1016/j.neucom.2010.03.021} {\bibfield  {journal} {\bibinfo
   {journal} {Neurocomputing}\ }\textbf {\bibinfo {volume} {74}},\ \bibinfo
  {pages} {239--255} (\bibinfo {year} {2010})},\ \bibinfo {note} {artificial
  Brains}\BibitemShut {NoStop}%
\bibitem [{\citenamefont {{Dibazar}}\ \emph {et~al.}(2006)\citenamefont
  {{Dibazar}}, \citenamefont {{Bangalore}}, \citenamefont {{Hyungook Park}},
  \citenamefont {{George}}, \citenamefont {{Yamada}},\ and\ \citenamefont
  {{Berger}}}]{Dibazar2006}%
  \BibitemOpen
  \bibfield  {author} {\bibinfo {author} {\bibfnamefont {A.~A.}\ \bibnamefont
  {{Dibazar}}}, \bibinfo {author} {\bibfnamefont {A.}~\bibnamefont
  {{Bangalore}}}, \bibinfo {author} {\bibnamefont {{Hyungook Park}}}, \bibinfo
  {author} {\bibfnamefont {S.}~\bibnamefont {{George}}}, \bibinfo {author}
  {\bibfnamefont {W.}~\bibnamefont {{Yamada}}}, \ and\ \bibinfo {author}
  {\bibfnamefont {T.~W.}\ \bibnamefont {{Berger}}},\ }\bibfield  {title}
  {\enquote {\bibinfo {title} {Hardware implementation of dynamic synapse
  neural networks for acoustic sound recognition},}\ }in\ \href {\doibase
  10.1109/IJCNN.2006.246949} {\emph {\bibinfo {booktitle} {The 2006 IEEE
  International Joint Conference on Neural Network Proceedings}}}\ (\bibinfo
  {year} {2006})\ pp.\ \bibinfo {pages} {2015--2022}\BibitemShut {NoStop}%
\bibitem [{\citenamefont {Soriano}\ \emph {et~al.}(2015)\citenamefont
  {Soriano}, \citenamefont {Ort{\'\i}n}, \citenamefont {Keuninckx},
  \citenamefont {Appeltant}, \citenamefont {Danckaert}, \citenamefont
  {Pesquera},\ and\ \citenamefont {van~der Sande}}]{Soriano2015}%
  \BibitemOpen
  \bibfield  {author} {\bibinfo {author} {\bibfnamefont {M.~C.}\ \bibnamefont
  {Soriano}}, \bibinfo {author} {\bibfnamefont {S.}~\bibnamefont {Ort{\'\i}n}},
  \bibinfo {author} {\bibfnamefont {L.}~\bibnamefont {Keuninckx}}, \bibinfo
  {author} {\bibfnamefont {L.}~\bibnamefont {Appeltant}}, \bibinfo {author}
  {\bibfnamefont {J.}~\bibnamefont {Danckaert}}, \bibinfo {author}
  {\bibfnamefont {L.}~\bibnamefont {Pesquera}}, \ and\ \bibinfo {author}
  {\bibfnamefont {G.}~\bibnamefont {van~der Sande}},\ }\bibfield  {title}
  {\enquote {\bibinfo {title} {Delay-based reservoir computing: noise effects
  in a combined analog and digital implementation},}\ }\href {\doibase
  https://doi.org/10.1109/TNNLS.2014.2311855} {\bibfield  {journal} {\bibinfo
  {journal} {IEEE transactions on neural networks and learning systems}\
  }\textbf {\bibinfo {volume} {26}},\ \bibinfo {pages} {388--393} (\bibinfo
  {year} {2015})}\BibitemShut {NoStop}%
\bibitem [{\citenamefont {Frye}, \citenamefont {Rietman},\ and\ \citenamefont
  {Wong}(1991)}]{Frye1991}%
  \BibitemOpen
  \bibfield  {author} {\bibinfo {author} {\bibfnamefont {R.}~\bibnamefont
  {Frye}}, \bibinfo {author} {\bibfnamefont {E.}~\bibnamefont {Rietman}}, \
  and\ \bibinfo {author} {\bibfnamefont {C.}~\bibnamefont {Wong}},\ }\bibfield
  {title} {\enquote {\bibinfo {title} {Back-propagation learning and
  nonidealities in analog neural network hardware},}\ }\href {\doibase
  10.1109/72.80296} {\bibfield  {journal} {\bibinfo  {journal} {IEEE
  Transactions on Neural Networks}\ }\textbf {\bibinfo {volume} {2}},\ \bibinfo
  {pages} {110--117} (\bibinfo {year} {1991})}\BibitemShut {NoStop}%
\bibitem [{\citenamefont {Semenova}\ \emph {et~al.}(2019)\citenamefont
  {Semenova}, \citenamefont {Porte}, \citenamefont {Andreoli}, \citenamefont
  {Jacquot}, \citenamefont {Larger},\ and\ \citenamefont
  {Brunner}}]{Semenova2019}%
  \BibitemOpen
  \bibfield  {author} {\bibinfo {author} {\bibfnamefont {N.}~\bibnamefont
  {Semenova}}, \bibinfo {author} {\bibfnamefont {X.}~\bibnamefont {Porte}},
  \bibinfo {author} {\bibfnamefont {L.}~\bibnamefont {Andreoli}}, \bibinfo
  {author} {\bibfnamefont {M.}~\bibnamefont {Jacquot}}, \bibinfo {author}
  {\bibfnamefont {L.}~\bibnamefont {Larger}}, \ and\ \bibinfo {author}
  {\bibfnamefont {D.}~\bibnamefont {Brunner}},\ }\bibfield  {title} {\enquote
  {\bibinfo {title} {Fundamental aspects of noise in analog-hardware neural
  networks},}\ }\href {\doibase 10.1063/1.5120824} {\bibfield  {journal}
  {\bibinfo  {journal} {Chaos: An Interdisciplinary Journal of Nonlinear
  Science}\ }\textbf {\bibinfo {volume} {29}},\ \bibinfo {pages} {103128}
  (\bibinfo {year} {2019})},\ \Eprint
  {http://arxiv.org/abs/https://doi.org/10.1063/1.5120824}
  {https://doi.org/10.1063/1.5120824} \BibitemShut {NoStop}%
\bibitem [{\citenamefont {Semenova}, \citenamefont {Larger},\ and\
  \citenamefont {Brunner}(2022)}]{Semenova2022}%
  \BibitemOpen
  \bibfield  {author} {\bibinfo {author} {\bibfnamefont {N.}~\bibnamefont
  {Semenova}}, \bibinfo {author} {\bibfnamefont {L.}~\bibnamefont {Larger}}, \
  and\ \bibinfo {author} {\bibfnamefont {D.}~\bibnamefont {Brunner}},\
  }\bibfield  {title} {\enquote {\bibinfo {title} {Understanding and mitigating
  noise in trained deep neural networks},}\ }\href {\doibase
  https://doi.org/10.1016/j.neunet.2021.11.008} {\bibfield  {journal} {\bibinfo
   {journal} {Neural Networks}\ }\textbf {\bibinfo {volume} {146}},\ \bibinfo
  {pages} {151--160} (\bibinfo {year} {2022})}\BibitemShut {NoStop}%
\bibitem [{\citenamefont {Everitt}(1998)}]{Everitt1998}%
  \BibitemOpen
  \bibfield  {author} {\bibinfo {author} {\bibfnamefont {B.}~\bibnamefont
  {Everitt}},\ }\href@noop {} {\emph {\bibinfo {title} {The Cambridge
  Dictionary of Statistics}}}\ (\bibinfo  {publisher} {Cambridge University
  Press},\ \bibinfo {address} {Cambridge, UK New York},\ \bibinfo {year}
  {1998})\BibitemShut {NoStop}%
\bibitem [{\citenamefont {LeCun}(2021)}]{LeCunSite}%
  \BibitemOpen
  \bibfield  {author} {\bibinfo {author} {\bibfnamefont {Y.}~\bibnamefont
  {LeCun}},\ }\href@noop {} {}\bibinfo {howpublished}
  {\url{http://yann.lecun.com/exdb/mnist/index.html}} (\bibinfo {year}
  {2021})\BibitemShut {NoStop}%
\bibitem [{\citenamefont {Chollet}\ \emph {et~al.}(2015)\citenamefont {Chollet}
  \emph {et~al.}}]{Keras}%
  \BibitemOpen
  \bibfield  {author} {\bibinfo {author} {\bibfnamefont {F.}~\bibnamefont
  {Chollet}} \emph {et~al.},\ }\bibfield  {title} {\enquote {\bibinfo {title}
  {Keras},}\ }\href@noop {} {\bibfield  {journal} {\bibinfo  {journal}
  {GitHub}\ } (\bibinfo {year} {2015})},\ \Eprint
  {http://arxiv.org/abs/https://github.com/fchollet/keras}
  {https://github.com/fchollet/keras} \BibitemShut {NoStop}%
\bibitem [{\citenamefont {Montgomery}\ and\ \citenamefont
  {Runger}(2002)}]{Montgomery2002}%
  \BibitemOpen
  \bibfield  {author} {\bibinfo {author} {\bibfnamefont {D.~C.}\ \bibnamefont
  {Montgomery}}\ and\ \bibinfo {author} {\bibfnamefont {G.~C.}\ \bibnamefont
  {Runger}},\ }\href@noop {} {\emph {\bibinfo {title} {Applied Statistics and
  Probability for Engineers -- 3rd ed.}}}\ (\bibinfo  {publisher} {John Wiley
  \and Sons},\ \bibinfo {year} {2002})\BibitemShut {NoStop}%
\end{thebibliography}

%

\end{document}